\documentclass{article}

\usepackage{microtype}
\usepackage{graphicx}
\usepackage{subfigure}
\usepackage{booktabs} 

\usepackage{hyperref}
\usepackage[in]{fullpage}  
\usepackage{mathpazo} 

\usepackage{algorithm}
\usepackage{amsmath}
\usepackage{array}
\usepackage{color}
\usepackage{comment}
\usepackage{fancybox}
\usepackage{float}
\usepackage[T1]{fontenc}
\usepackage{gensymb}
\usepackage{multirow}
\usepackage{natbib}
\usepackage{pgfplots}
\usepackage{tikz}
\usepackage[textsize=scriptsize,textwidth=3cm]{todonotes} 
\usepackage{wrapfig}

\newcommand\eps{\epsilon}

\renewcommand\bar\overline
\renewcommand\epsilon\varepsilon
\newcommand\px{\textrm{px}}

\newcommand\AND{
    \end{tabular}\hfil\linebreak[4]\hfil%
    \begin{tabular}[t]{c}\ignorespaces%
}

\title{Exploring the Landscape of Spatial Robustness}

\author{
Logan Engstrom\thanks{Equal Contribution}\\
MIT\\
{\tt engstrom@mit.edu}
    \and
Brandon Tran\footnotemark[1]\\
MIT\\
{\tt btran115@mit.edu}
    \and
Dimitris Tsipras\footnotemark[1]\\
MIT\\
{\tt tsipras@mit.edu}
    \AND
Ludwig Schmidt\\
MIT\\
{\tt ludwigs@mit.edu}
    \and
Aleksander M\k{a}dry\\
MIT\\
{\tt madry@mit.edu}
}
\date{}

\begin{document}

\maketitle

\begin{abstract}
The study of adversarial robustness has so far largely focused on perturbations
bound in $\ell_p$-norms.
However, state-of-the-art models turn out to be also vulnerable to other, more
natural classes of perturbations such as translations and rotations.
In this work, we thoroughly investigate the vulnerability of
neural network--based
classifiers to rotations and translations. While data
augmentation offers relatively small robustness, we use
ideas from robust
optimization and test-time input aggregation to significantly improve
robustness.
Finally we find that, in contrast to the $\ell_p$-norm case,
first-order
methods cannot reliably find worst-case perturbations. This highlights
spatial robustness as a fundamentally different setting requiring additional
study.\footnote{Code for CIFAR10 available at
\url{https://github.com/MadryLab/adversarial_spatial} (TensorFlow) and for
ImageNet at \url{https://github.com/MadryLab/spatial-pytorch} (PyTorch). }

\end{abstract}

\section{Introduction}
\label{sec:intro}
Neural networks are now widely embraced as dominant solutions in computer
vision~\citep{krizhevsky2012imagenet,he2016deep}, speech
recognition~\citep{graves2013speech}, and natural language
processing~\citep{collobert2008unified}. However, while their accuracy scores
often match (and sometimes go beyond) human-level performance on key
benchmarks~\citep{he2015delving,taigman2014deepface}, models experience severe performance
degradation in the worst case. 

In this case, models are vulnerable to so-called \emph{adversarial examples}, or
slightly perturbed inputs that are almost indistinguishable from natural data to
a human but cause state-of-the-art classifiers to make incorrect
predictions~\citep{szegedy2014intriguing,goodfellow2015explaining}. This raises concerns about
the use of neural networks in contexts where reliability, dependability, and
security are important.

There is a long line of work on methods for constructing adversarial
perturbations in various
settings~\citep{szegedy2014intriguing,goodfellow2015explaining,kurakin2017adversarial,sharif2016accessorize,dezfooli2016deepfool,carlini2017towards,papernot2017practical,madry2018towards,athalye2018synthesizing}.
However, these methods largely rely on perturbations that are quite contrived
and hence unlikely to exist in the real world.
For example, in the context of image recognition, a large body of work has
focused on fooling models with carefully crafted, $\ell_p$-bounded
perturbations in pixel-space.

As such, one may suspect that adversarial examples are a problem only in the
presence of a truly malicious attacker, and are unlikely to arise in more benign
environments. However, recent work has shown that neural network--based vision
classifiers are vulnerable to input images that have been \textit{spatially
  transformed} through small rotations, translations, shearing, scaling, and
other natural transformations~\citep{fawzi2015manitest, kanbak2018geometric,
  xiao2018spatially, boneh2017personal}.
Such transformations are pervasive in vision applications and hence quite likely
to naturally occur in practice.
The vulnerability of neural networks to
such transformations raises a natural question:

\begin{center}
  \emph{How can we build spatially robust classifiers?}
\end{center}

We address this question by first performing an in-depth study of neural
network--based classifier robustness to two basic image transformations:
translations and rotations. While these transformations appear natural to a
human, we show that small rotations and translations \emph{alone} can
significantly degrade accuracy. These transformations are particularly relevant
for computer vision applications since real-world objects do not always appear
perfectly centered.

\begin{figure}[!htp]
\begin{center}
{\setlength\tabcolsep{0pt}
\begin{tabular}{cccccc}
Natural & Adversarial & Natural & Adversarial & Natural & Adversarial\\
\includegraphics[width=0.16\textwidth]{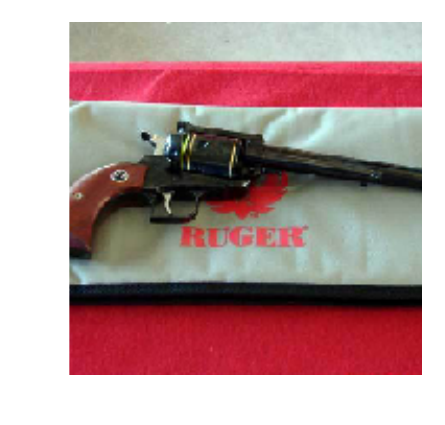} &
\includegraphics[width=0.16\textwidth]{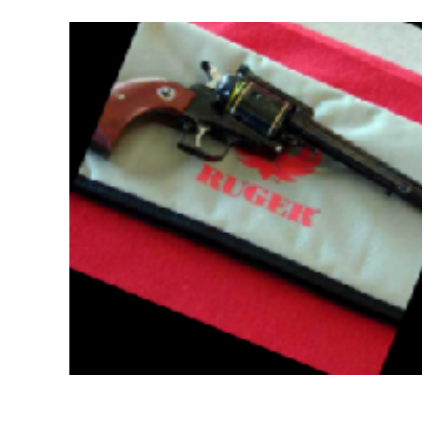} &
\includegraphics[width=0.16\textwidth]{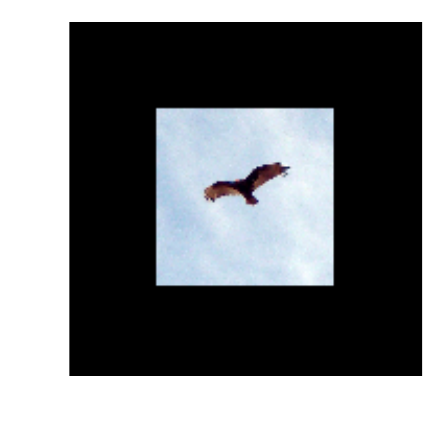} &
\includegraphics[width=0.16\textwidth]{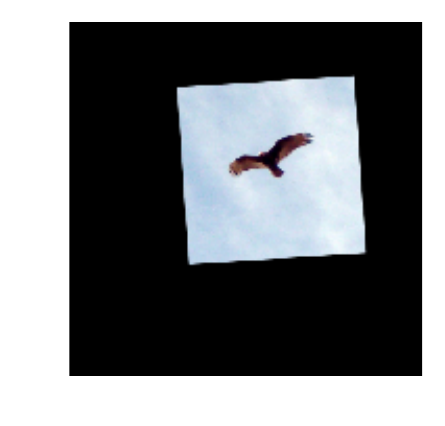} &
\includegraphics[width=0.16\textwidth]{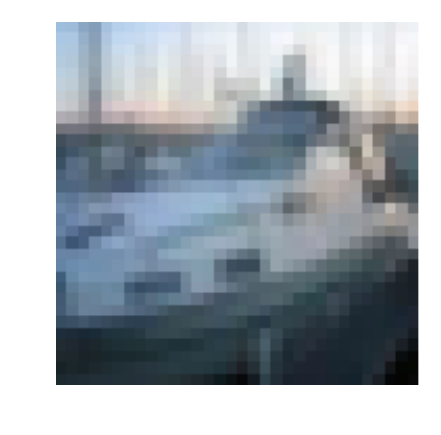} &
\includegraphics[width=0.16\textwidth]{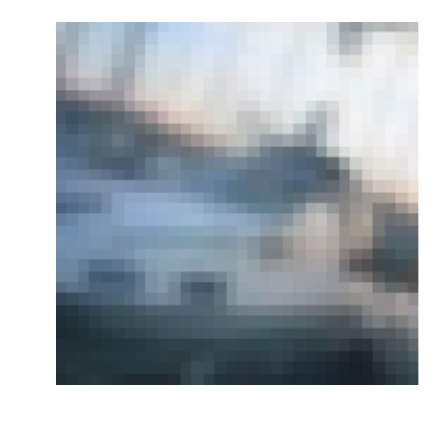} 
\vspace{-5pt} \\
``revolver'' & ``mousetrap'' & ``vulture'' & ``orangutan'' & ``ship'' & ``dog''
\end{tabular}
}
\end{center}
\caption{Examples of adversarial transformations and their predictions in the standard, "black canvas", and reflection padding setting.}
\label{fig:small_fool}
\end{figure}

\subsection{Our Methodology and Results}
Our goal is to obtain a fine-grained understanding of the spatial robustness of
standard, near state-of-the-art image classifiers for the
MNIST~\citep{lecun1998mnist}, CIFAR10~\citep{krizhevsky2009learning}, and
ImageNet~\citep{russakovsky2015imagenet} datasets.

\vspace{-.3cm}
\paragraph{Classifier brittleness.}
We find that small rotations and translations consistently and
significantly degrade accuracy of image classifiers on a number of tasks, as
illustrated in Figure~\ref{fig:small_fool}.
Our results suggest that classifiers are highly brittle: even small \emph{random} transformations can degrade accuracy by up to 30\%.
Such brittleness to random transformations suggests that these models might be
unreliable even in benign settings.

\vspace{-.3cm}
\paragraph{Relative adversary strength.}
We then perform a thorough analysis comparing the abilities of various
adversaries---first-order, random, and grid-based---to fool models with small
rotations and translations. 
In particular, we find that exhaustive grid search-based adversaries are much
more powerful than first-order adversaries.
This is in stark contrast to results in the $\ell_p$-bounded adversarial example
literature, where first-order methods can consistently find approximately
worst-case inputs~\citep{carlini2017towards, madry2018towards}. 

\vspace{-.3cm}
\paragraph{Spatial loss landscape.}
To understand why such a difference occurs, we delve deeper into the classifiers
to try and understand the failure modes induced by such natural transformations.
We find that the loss landscape of classifiers with respect to rotations and
translations is highly non-concave and contains many spurious maxima.
This is, again, in contrast to the $\ell_p$-bounded setting, in which,
experimentally, the value of different maxima tend to concentrate
well~\citep{madry2018towards}.
Our loss landscape results thus demonstrate that any adversary relying on first
order information might be unable to reliably find misclassifications.
Consequently, rigorous evaluation of model robustness in this spatial setting
requires techniques that that go beyond what was needed to induce $\ell_p$-based
adversarial robustness. 

\vspace{-.3cm}
\paragraph{Improving spatial robustness.}
We then develop methods for alleviating these vulnerabilities using insights
from our study.
As a natural baseline, we augment the training procedure with rotations and
translations.
While this does largely mitigate the problem on MNIST, additional data
augmentation only marginally increases robustness on CIFAR10 and ImageNet.
We thus propose two natural methods for further increasing the robustness of
these models, based on robust optimization and aggregation of random input
transformations. 
These methods offer significant improvements in classification accuracy against
both adaptive and random attackers when compared to both standard models and
those trained with additional data augmentation. 
In particular, on ImageNet, our best model attains a top1 accuracy of
\textbf{56\%} against the strongest adversary, versus \textbf{34\%} for a
standard network with additional data augmentation. 

\vspace{-.3cm}
\paragraph{Combining spatial and $\ell_\infty$-bounded attacks.}
Finally, we examine the interplay between spatial and
$\ell_\infty$-based perturbations.
We observe that robustness to these two classes of input perturbations is
largely orthogonal.
In particular, pixel-based robustness does not imply spatial robustness, while
combining spatial and $\ell_\infty$-bounded transformations seems to have a
\emph{cumulative} effect in reducing classification accuracy.
This emphasizes the need to broaden the notions of image similarity in the
adversarial examples literature beyond the common $\ell_p$-balls.

\section{Related Work}
The fact that small rotations and translation can fool neural networks on MNIST
and CIFAR10 was, to the best of our knowledge, first observed in~\citep{fawzi2015manitest}.
They compute the minimum transformation required to fool the model and use it as a measure for a quantitative comparison of different architectures and training procedures.
The main difference to our work is that we focus on the optimization aspect of the problem.
We show that a few random queries usually suffice for a successful attack, while first-order methods are ineffective.
Moreover, we go beyond standard data augmentation and evaluate the effectiveness of natural baseline defenses.

The concurrent work of \cite{kanbak2018geometric} proposes a different first-order method to evaluate the robustness of classifiers based on geodesic distances on a manifold.
This metric is harder to interpret than our parametrized attack space.
Moreover, given our findings on the non-concavity of the optimization landscape, it is unclear how close their method is to the ground truth (exhaustive enumeration).
While they perform a limited study of defenses (adversarial fine-tuning) using their method, it appears to be less effective than our baseline worst-of-10 training.
We attribute this difference to the inherent obstacles first-order methods face in this optimization landscape.

Recently, \cite{xiao2018spatially} and \cite{boneh2017personal} observed
independently that it is possible to use various spatial transformations to
construct adversarial examples for naturally and $\ell_\infty$-adversarially trained models.
The main difference from our work is that we show even very simple transformations (translations and rotations) are sufficient to break a variety of classifiers, while the transformations employed in~\citep{xiao2018spatially} and \citep{boneh2017personal} are more involved.
The transformation in~\citep{xiao2018spatially} is based on performing a displacement of individual pixels in the original image constrained to be globally smooth and then optimized for misclassification probability.
\cite{boneh2017personal} consider an $\ell_\infty$-bounded pixel-wise perturbation of a version of the original image that has been slightly rotated and in which a few random pixels have been flipped.
Both of these methods require direct access to the attacked model (or a surrogate) to compute (or at least estimate) the gradient of the loss function with respect to the model's input. 
In contrast, our attacks can be implemented using only a small number of random,
non-adaptive inputs.

\section{Adversarial Rotations and Translations}
\label{sec:attacks}
Recall that in the context of image classification, an \emph{adversarial example} for a given input image $x$ and a classifier $C$ is an image $x'$ that satisfies two properties:
(i) on the one hand, the adversarial example $x'$ causes the classifier $C$ to output a different label on $x'$ than on $x$, i.e., we have $C(x)\neq C(x')$.
(ii) On the other hand, the adversarial example $x'$ is ``visually similar'' to $x$.

Clearly, the notion of visual similarity is not precisely defined here.
In fact, providing a precise and rigorous definition is extraordinarily difficult as it would require formally capturing the notion of human perception.
Consequently, previous work largely settled on the assumption that $x'$ is a valid adversarial example for $x$ if and only if $\|x - x'\|_p\leq \eps$ for some $p\in[0,\infty]$ and $\eps$ small enough.
This convention is based on the fact that two images are indeed visually similar when they are close enough in some $\ell_p$-norm.
However, the converse is not necessarily true.
A small rotation or translation of an image usually appears visually similar to a human, yet can lead to a large change when measured in an $\ell_p$-norm.
We aim to expand the range of similarity measures considered in the adversarial examples literature by investigating robustness to small rotations and translations.

\paragraph{Attack methods.} Our first goal is to develop sufficiently strong methods for generating adversarial rotations and translations.
In the context of pixel-wise $\ell_p$-bounded perturbations, the most successful
approach for constructing adversarial examples so far has been to employ
optimization methods on a suitable loss function
\citep{szegedy2014intriguing,goodfellow2015explaining,carlini2017towards}.
Following this approach, we parametrize our attack method with a set of tunable parameters and then optimize over these parameters.

First, we define the exact range of attacks we want to optimize over.
For the case of rotation and translation attacks, we wish to find parameters $(\delta u, \delta v, \theta)$ such that rotating the original image by $\theta$ degrees around the center and then translating it by $(\delta u, \delta v)$ pixels causes the classifier to make a wrong prediction.
Formally, the pixel at position $(u, v)$ is moved to the following position (assuming the point $(0,0)$ is the center of the image):
$$\begin{bmatrix} u' \\ v'\end{bmatrix} = \begin{bmatrix}
  \cos\theta & -\sin\theta \\ \sin\theta & \cos\theta \end{bmatrix}
  \cdot\begin{bmatrix} u \\ v\end{bmatrix}
  +\begin{bmatrix} \delta u \\ \delta v\end{bmatrix}.$$
We implement this transformation in a differentiable manner using the spatial
transformer blocks of~\citep{jaderberg2015spatial}~\footnote{We used the open
  source implementation found here: \url{https://github.com/tensorflow/models/tree/master/research/transformer}.}.
In order to handle pixels that are mapped to non-integer coordinates, the
transformer units include a differentiable bilinear interpolation routine.
Since our loss function is differentiable with respect to the input and the
transformation is in turn differentiable with respect to its parameters, we can obtain gradients of the model's loss function w.r.t.\ the perturbation parameters.
This enables us to apply a first-order optimization method to our problem. 

By defining the spatial transformation for some $x$ as $T(x; \delta u, \delta
v, \theta)$, we construct an adversarial perturbation for $x$ by
solving the problem
\begin{equation}
\label{eq:optimization}
\max_{\delta u, \delta v, \theta}
  \mathcal{L}(x',y),\quad\textrm{for}\ x' = T(x; \delta u, \delta v, \theta) \; ,
\end{equation}
where $\mathcal{L}$ is the loss function of the neural network\footnote{The loss
$\mathcal{L}$ of the classifier is a function from images to real numbers that
expresses the performance of the network on the particular example $x$ (e.g., the cross-entropy between predicted and correct distributions).}, and $y$ is the
correct label for $x$.

We compute the perturbation from Equation~\ref{eq:optimization} in three distinct ways:
\begin{itemize}
\item {\bf First-Order Method (FO):}
Starting from a random choice of parameters, we iteratively take steps in the direction of the gradient of the loss function.
This is the direction that locally maximizes the loss of the classifier (as a surrogate for misclassification probability).
Since the maximization problem we are optimizing is non-concave, there are no guarantees for global optimality, but the hope is that the local maximum solution closely approximates the global optimum.
Note that unlike the $\ell_p$-norm case, we are not optimizing in the pixel
space but in the latent space of rotation and translation parameters.

\item {\bf Grid Search:} We discretize the parameter space and exhaustively examine every possible parametrization of the attack to find one that causes the classifier to give a wrong prediction (if such a parametrization exists).
  Since our parameter space is low-dimensional enough, this method is computationally feasible (in contrast to a grid search for $\ell_p$-based adversaries).

\item {\bf Worst-of-$k$:} We randomly sample $k$ different choices of attack parameters and choose the one on which the model performs worst.
As we increase $k$, this attack interpolates between a random choice and grid search.
\end{itemize}

We remark that while a first-order attack requires full knowledge of the model to compute the gradient of the loss with respect to the input, the other two attacks do not.
They only require the outputs corresponding to chosen inputs, which can be done with only query access to the target model.

\section{Improving Invariance to Spatial Transformations}
\begin{figure*}[!htp]
  \centering
\includegraphics[width=.85\textwidth]{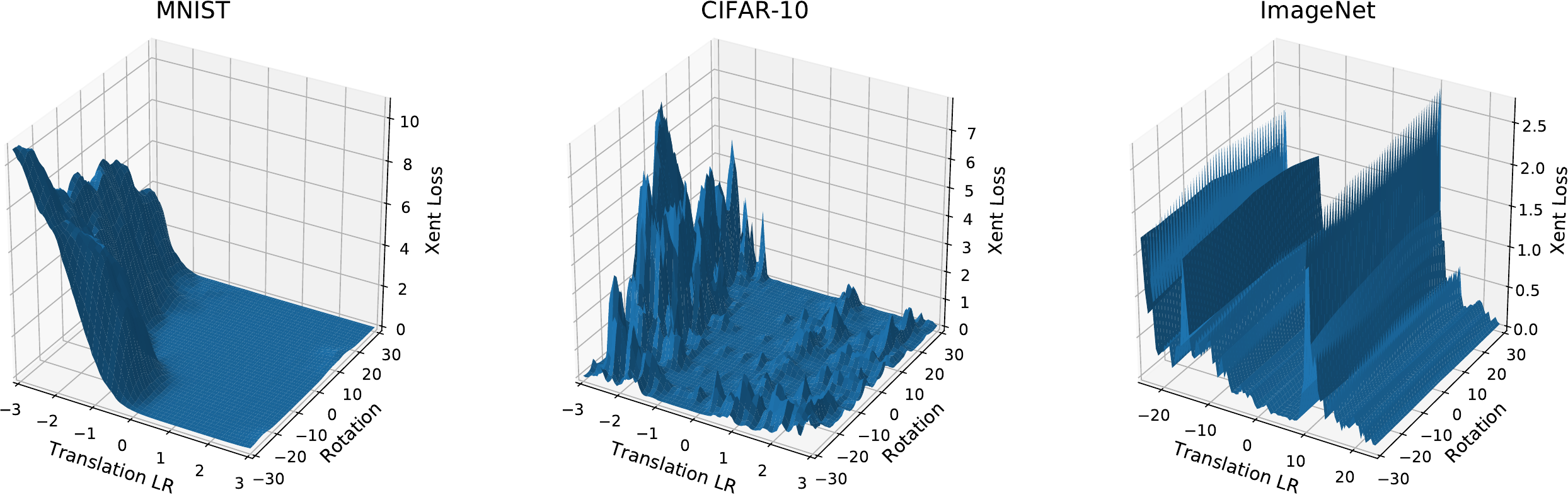}
\caption{Loss landscape of a random example for each dataset when performing left-right translations and rotations.
Translations and rotations are restricted to 10\% of the image pixels and $30\degree$, respectively.
We observe that the landscape is significantly non-concave, rendering first-order methods to generate adversarial example ineffective.
Figure~\ref{fig:landscape_complete} in the appendix shows additional examples.}
\label{fig:landscape}
\end{figure*}
As we will see in Section~\ref{sec:exp}, augmenting the training set with random rotations and translations does improve the robustness of the model against such random transformations.
However, data augmentation does not significantly improve the robustness against worst-case attacks and sometimes leads to a drop in accuracy on unperturbed images.
To address these issues, we explore two simple baselines that turn out to be surprisingly effective.

\paragraph{Robust Optimization.}
Instead of performing standard empirical risk minimization to train the classification model, we utilize ideas from robust optimization.
Robust optimization has a rich history~\citep{ben2009robust} and has recently been
applied successfully in the context of defending neural networks against
adversarial
examples~\citep{madry2018towards,sinha2018certifiable,raghunathan2018certified,wong2018provable}.
The main barrier to applying robust optimization for spatial transformations is the lack of an efficient procedure to compute the worst-case perturbation of a given example.
Performing a grid search (as described in Section~\ref{sec:attacks}) is prohibitive as this would increase the training time by a factor close to the grid size, which can easily be a factor 100 or 1,000.
Moreover, the non-convexity of the loss landscape prevents potentially more efficient first-order methods from discovering (approximately) worst-case transformations (see Section~\ref{sec:exp} for details).

Given that we cannot fully optimize over the space of translations and rotations, we instead use a coarse approximation provided by the worst-of-10 adversary (as described in Section~\ref{sec:attacks}).
So each time we use an example during training, we first sample 10 transformations of the example uniformly at random from the space of allowed transformations.
We then evaluate the model on each of these transformations and train on the one perturbation with the highest loss.
This corresponds to approximately minimizing a min-max formulation of robust
accuracy similar to~\citep{madry2018towards}.
Training against such an adversary increases the overall time by a factor of roughly six.\footnote{We need to perform 10 forward passes and one backwards pass instead of one forward and one backward pass required for standard training.}

\paragraph{Aggregating Random Transformations.}
As Section~\ref{sec:exp} shows, the accuracy against a \emph{random} transformation is significantly higher than the accuracy against the worst transformation in the allowed attack space.
This motivates the following inference procedure: compute a (typically small)
number of random transformations of the input image and output the label that
occurs the most in the resulting set of predictions.
We constrain these random transformations to be within $5\%$ of the input image size in each translation direction and up to $15\degree$ of rotation.
\footnote{Note that if an adversary rotates an image by $30\degree$ (a valid attack in our threat model), we may end up evaluating the image on rotations of up to $45\degree$.}
The training procedure and model can remain unchanged while the inference time is increased by a small factor (equal to the number of transformations we evaluate on).

\paragraph{Combining Both Methods.}
The two methods outlined above are orthogonal and in some sense complementary.
We can therefore combine robust training (using a worst-of-k adversary) and majority inference to further increase the robustness of our models.

\section{Experiments}
\label{sec:exp}
\begin{figure*}[!htp]
\begin{center}
\begin{tabular}{ccc}
\includegraphics[scale=0.27]{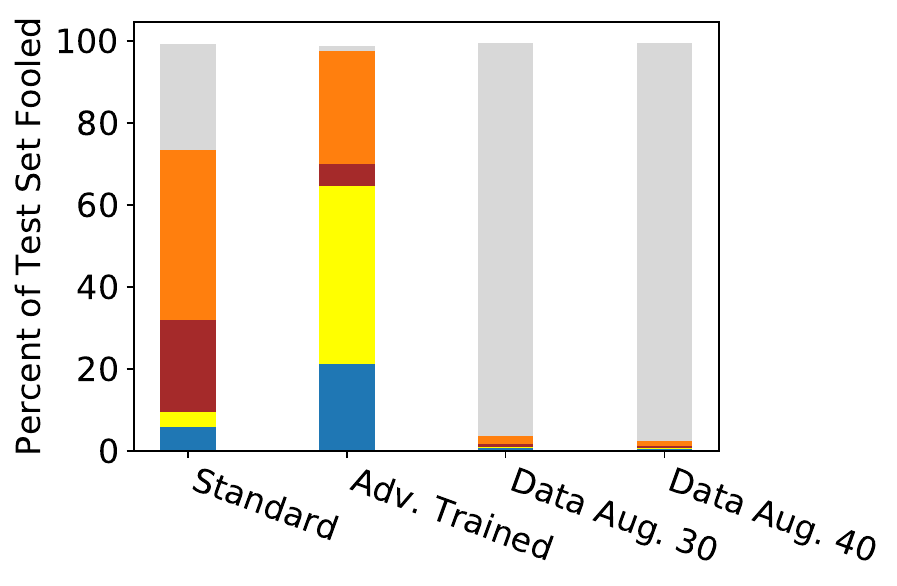} &
\includegraphics[scale=0.27]{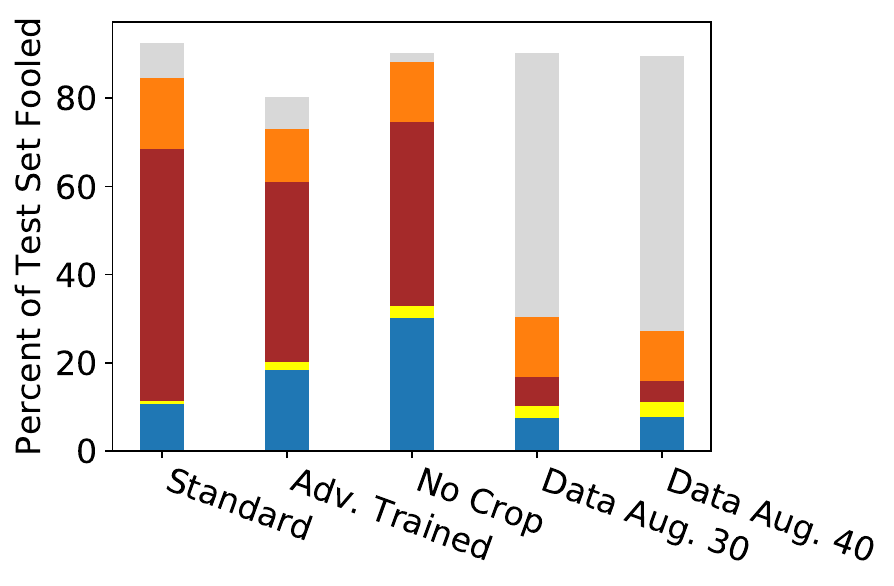} &
\includegraphics[scale=0.27]{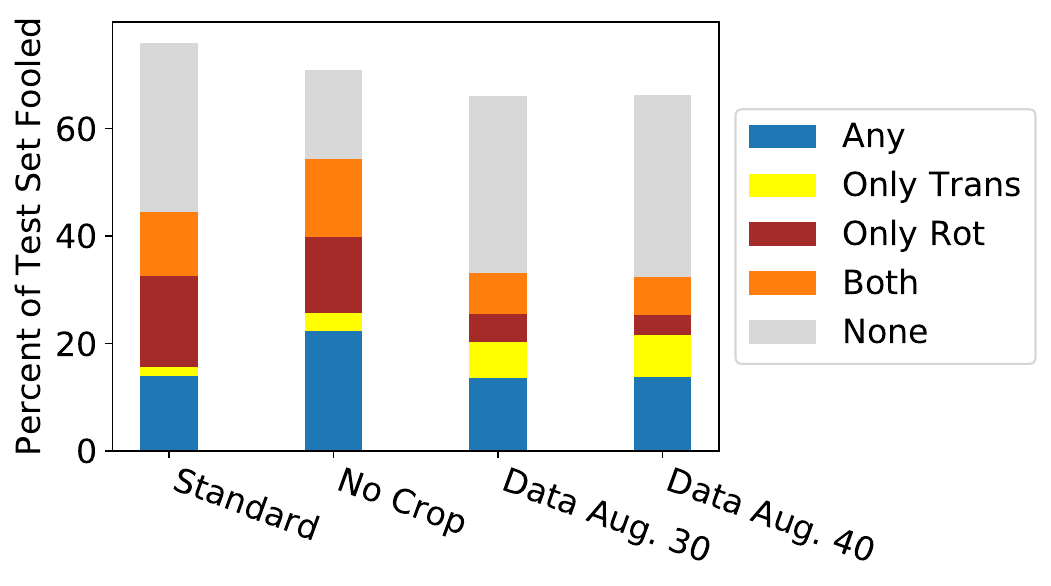} \\
MNIST & CIFAR10 & \hspace{-1cm}ImageNet
\end{tabular}
\caption{Fine-grained dataset analysis. For each model, we visualize what percent of the test set can be fooled via various methods. We compute how many examples can be fooled with either translations or rotations ("any"), how many can be fooled only by one of these, and how many require a combination to be fooled ("both").}
\label{fig:bars}
\end{center}
\end{figure*}

We evaluate standard image classifiers for the MNIST~\citep{lecun1998mnist},
CIFAR10~\citep{krizhevsky2009learning} and
ImageNet~\citep{russakovsky2015imagenet} datasets.
In order to determine the extent to which misclassification is caused by insufficient data augmentation during training, we examine various data augmentation methods.
We begin with a description of our experimental setup.

\paragraph{Model Architecture.} For MNIST, we use a convolutional neural network derived from the TensorFlow Tutorial~\citep{abadi2016tensorflow}.
In order to obtain a fully convolutional version of the network, we replace the fully-connected layer by two convolutional layers with 128 and 256 filters each, followed by a global average pooling.
For CIFAR10, we consider a standard ResNet~\citep{he2016deep} model with 4 groups of residual layers with filter sizes [16, 16, 32, 64] and 5 residual units each.
We use standard and $\ell_\infty$-adversarially trained models similar to those
studied by \cite{madry2018towards}.\footnote{\url{https://github.com/MadryLab/cifar10_challenge}}\textsuperscript{,}\footnote{\url{https://github.com/MadryLab/mnist_challenge}}
For ImageNet, we use a ResNet-50~\citep{he2016deep} architecture implemented in
the \texttt{tensorpack} repository~\citep{wu2016tensorpack}.
We did not modify the model architectures or training procedures.

\paragraph{Attack Space.} In order to maintain the visual similarity of images to the natural ones we restrict the space of allowed perturbations to be relatively small.
We consider rotations of at most $30\degree$ and translations of at most (roughly) 10\% percent of the image size in each direction.
This corresponds to 3 pixels for MNIST (image size $28\times28$) and CIFAR10 (image size $32\times32$), and 24 pixels for ImageNet (image size $299\times299$).
For grid search attacks, we consider 5 values per translation direction and 31 values for rotations, equally spaced. 
For first-order attacks, we use $200$ steps of projected gradient descent of step size $0.01$ times the parameter range.
When rotating and translating the images, we fill the empty space with zeros (black pixels).

\paragraph{Data Augmentation.} We consider five variants of training for our models.
\begin{itemize}
\item Standard training: The standard training procedure for the respective model architecture.

\item $\ell_\infty$-bounded adversarial training: The classifier is trained on $\ell_\infty$-bounded adversarial examples that are generated with projected gradient descent.

\item No random cropping: Standard training for CIFAR-10 and ImageNet includes data augmentation via random crops. We investigate the effect of this data augmentation scheme by also training a model without random crops.

\item Random rotations and translations: At each training step, we perform a uniformly random perturbation from the attack space on each training example.

\item Random rotations and translations from larger intervals: As before, we perform uniformly random perturbations, but now from a \emph{superset} of the attack space ($40\degree$, $\pm$ 13\% pixels).

\end{itemize}

\subsection{Evaluating Model Robustness}
We evaluate all models against random and grid search adversaries with rotations and translations considered both separately and together.
We report the results in Table~\ref{tab:core}.
We visualize a random subset of successful attacks in Figures~\ref{fig:mnist_fool},~\ref{fig:cifar_fool}, and~\ref{fig:imagenet_fool} of Appendix~\ref{app:omitted}.

Despite the high accuracy of standard models on unperturbed examples and their reasonable performance on random perturbations, a grid search can significantly lower the classifiers' accuracy on the test set.
For the standard models, accuracy drops from 99\% to 26\% on MNIST, 93\% to 3\% on CIFAR10, and 76\% to 31\% on ImageNet (Top~1 accuracy).

The addition of random rotations and translations during training greatly improves both the random and adversarial accuracy of the classifier for MNIST and CIFAR10, but less so for ImageNet.
For the first two datasets, data augmentation increases the accuracy against a grid adversary by 60\% to 70\%, while the same data augmentation technique adds less than 3\% accuracy on ImageNet.

We perform a fine-grained investigation of our findings:
\begin{itemize}
\item In Figure~\ref{fig:bars} we examine how many examples can be fooled by (i) rotations only, (ii) translations only, (iii) neither transformation, or (iv) both.

\item We visualize the set of fooling angles for a random sample of the
  rotations-only grid in Figure~\ref{fig:angles_main} on ImageNet, and provide
  more examples in the appendix in Figure~\ref{fig:angles}. We observe that the set of fooling angles is nonconvex and not contiguous.

\item To investigate how many transformations are adversarial per image, we analyze the percentage of misclassified grid points for each example in Figure~\ref{fig:hist}.
While the majority of images has only a small number of adversarial transformations, a significant fraction of images is fooled by 20\% or more of the transformations.
\end{itemize}

\begin{figure*}[!htp]
  \centering
\includegraphics[scale=.3]{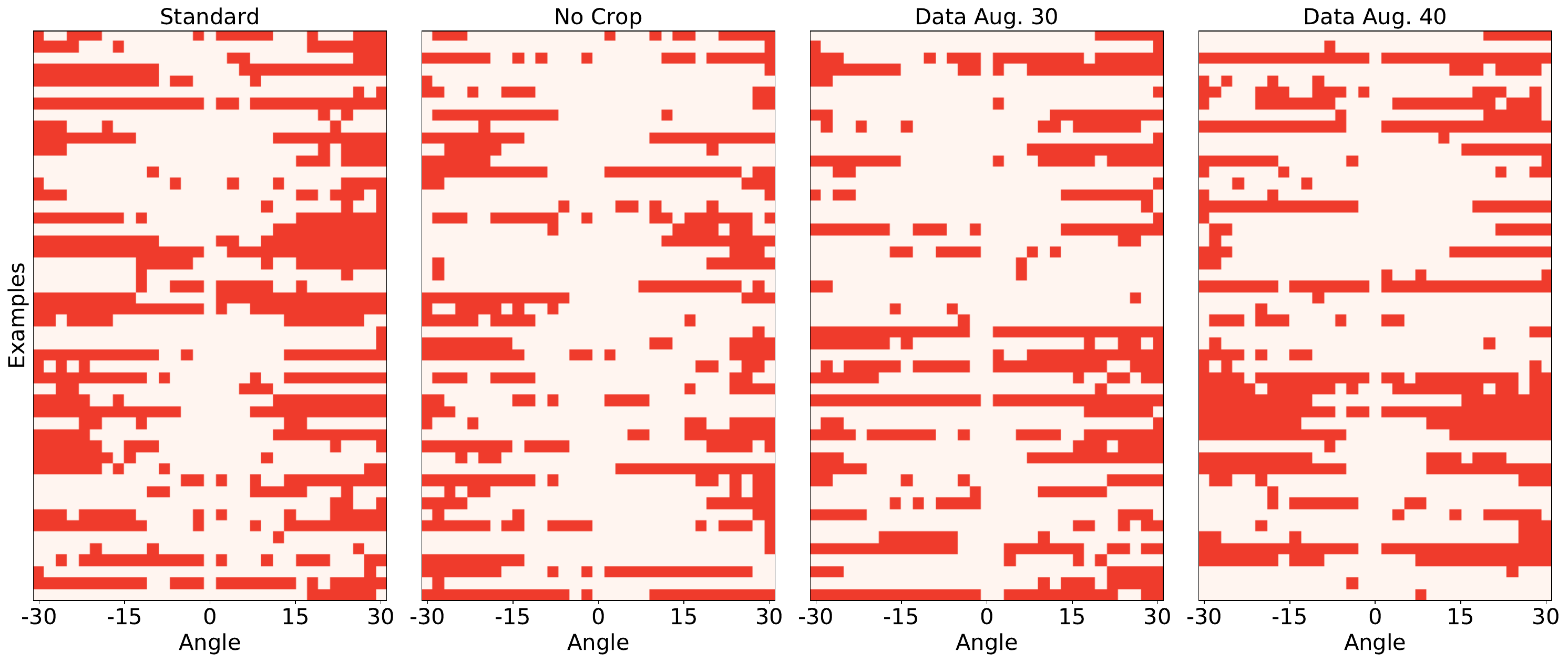}
\caption{Visualizing which angles fool ImageNet classifiers for 50 random examples.
  For each dataset and model, we visualize one example per row.
  \textcolor{red}{Red} corresponds to \emph{misclassification} of the images. We
  observe that the angles fooling the models form a highly non-convex set.
  Figure~\ref{fig:angles} in the appendix shows additional examples for CIFAR10
  and MNIST.}
\label{fig:angles_main}
\end{figure*}

\begin{table*}[!htp]
\caption{
Accuracy of different classifiers against rotation and translation adversaries on MNIST, CIFAR10, and ImageNet.
The allowed transformations are translations by (roughly) 10\% of the image size and $\pm30\degree$ rotations.
The attack parameters are chosen through random sampling or grid search with rotations and translations considered both together (``Rand.'', ``Grid'') and separately (``Rand. T.'' and ``Grid T.'' for transformations, ``Rand R.'' and ``Grid R.'' for rotations).
We consider networks that are trained with (i) the respective standard setup, (ii) no data augmentation (if data augmentation is present in standard setup), (iii) with an $\ell_\infty$ adversary, (iv) with data augmentation corresponding to the attack space ($\pm 3\px,\pm30\degree$) and an enlarged space ($\pm 4\px,\pm40\degree$), and (v) with worst-of-$10$ training for both types of augmentations.
}
\label{tab:mnist}
\label{tab:cifar}
\label{tab:imagenet}
\label{tab:core}
\begin{center}
{\renewcommand{\arraystretch}{1.1}
\begin{tabular}{c|c|ccc|cc|cc}
    & Model & Nat. & Rand. & Grid & Rand. T. & Grid T. & Rand. R. & Grid R.\\
\hline                                                                            
\multirow{6}{*}{\rotatebox[origin=c]{90}{MNIST}}
& Standard
    & 99.31\% & 94.23\% & \textbf{26.02}\% & 98.61\% & 89.80\% & 95.68\% & 70.98\% \\
& $\ell_\infty$-Adv
    & 98.65\% & 88.02\% & \textbf{ 1.20}\% & 93.72\% & 34.13\% & 95.27\% & 72.03\% \\
& Aug. 30
    & 99.53\% & 99.35\% & \textbf{95.79}\% & 99.47\% & 98.66\% & 99.34\% & 98.23\% \\
& Aug. 40 
    & 99.34\% & 99.31\% & \textbf{96.95}\% & 99.39\% & 98.65\% & 99.40\% & 98.49\% \\
& W-10 (30)
    & 99.48\% & 99.37\% & \textbf{97.32}\% & 99.50\% & 99.01\% & 99.39\% & 98.62\% \\
& W-10 (40)
    & 99.42\% & 99.39\% & \textbf{97.88}\% & 99.45\% & 98.89\% & 99.36\% & 98.85\% \\

\hline
\multirow{7}{*}{\rotatebox[origin=c]{90}{CIFAR10}}
& Standard
    & 92.62\% & 60.93\% & \textbf{ 2.80}\% & 88.54\% & 66.17\% & 75.36\% & 24.71\% \\
& No Crop
    & 90.34\% & 54.64\% & \textbf{ 1.86}\% & 81.95\% & 46.07\% & 69.23\% & 18.34\% \\
& $\ell_\infty$-Adv
    & 80.21\% & 58.33\% & \textbf{ 6.02}\% & 78.15\% & 59.02\% & 62.85\% & 20.98\% \\
& Aug. 30
    & 90.02\% & 90.92\% & \textbf{58.90}\% & 91.76\% & 79.01\% & 91.14\% & 76.33\% \\
& Aug. 40
    & 88.83\% & 91.18\% & \textbf{61.69}\% & 91.53\% & 77.42\% & 91.10\% & 76.80\% \\
& W-10 (30)
    & 91.34\% & 92.35\% & \textbf{69.17}\% & 92.43\% & 83.01\% & 92.33\% & 81.82\% \\
& W-10 (40)
    & 91.00\% & 92.11\% & \textbf{71.15}\% & 92.28\% & 82.15\% & 92.53\% & 82.25\% \\

\hline                                                                              
\multirow{5}{*}{\rotatebox[origin=c]{90}{ImageNet}}
& Standard
    & 75.96\%  & 63.39\% & \textbf{31.42}\% & 73.24\% &60.42\% & 67.90\%& 44.98\% \\
& No Crop                                                            
    & 70.81\%  & 59.09\% & \textbf{16.52}\% & 66.75\% &45.17\% & 62.78\%& 34.17\% \\
& Aug. 30                                 
    & 65.96\%  & 68.60\% & \textbf{32.90}\% & 70.27\% &45.72\% & 69.28\%& 47.25\% \\
& Aug. 40                                 
    & 66.19\%  & 67.58\% & \textbf{33.86}\% & 69.50\% &44.60\% & 68.88\%& 48.72\% \\
& W-10 (30)
    & 76.14\% & 73.19\% & \textbf{52.76}\% &74.42\% & 61.18\%  & 73.74\% & 61.06\% \\
& W-10 (40)
    & 74.64\% & 71.36\% & \textbf{50.23}\% &72.86\% & 59.34\%  & 71.95\% & 59.23\% \\
\end{tabular}}
\end{center}
\end{table*}

\paragraph{Padding Experiments.} A natural question is whether the reduced accuracy of the models is due to the cropping applied during the transformation.
We verify that this is not the case by applying zero and reflection padding to the image datasets.
We note that the zero padding creates a ``black canvas'' version of the dataset, ensuring that no information from the original image is lost after a transformation.
We show a random set of adversarial examples in this setting in Figure~\ref{fig:canvas_fool} and a full evaluation in Table~\ref{tab:canvas}.
We also provide more details regarding reflection padding in Section~\ref{app:reflection} and provide an evaluation in Table~\ref{tab:cifar_reflect}.
All of these are in Appendix~\ref{app:omitted}.

\subsection{Comparing Attack Methods}
In Table~\ref{tab:methods_small} we compare different attack methods on various classifiers and datasets.
We observe that worst-of-10 is a powerful adversary despite its limited interaction with the target classifier.
The first-order adversary performs significantly worse.
It fails to approximate the ground-truth accuracy of the models and performs significantly worse than the grid adversary and even the worst-of-10 adversary.
\begin{table*}[!htp]
\caption{
Comparison of attack methods across datasets and models. Worst-of-10 is very effective and significantly reduces the model accuracy despite the limited interaction. The first-order (FO) adversary performs poorly, despite the large number of steps allowed. We compare standard training to Augmentation $(\pm3\px,\pm30\degree)$.
For the full table, see Figure~\ref{tab:methods} of Appendix~\ref{app:omitted}.
}
\label{tab:methods_small}
\begin{center}
{\renewcommand{\arraystretch}{1.1}
\begin{tabular}{c|cc|cc|cc}
& \multicolumn{2}{c|}{MNIST}& \multicolumn{2}{c|}{CIFAR-10}& \multicolumn{2}{c}{ImageNet}\\
& Standard & Aug. & Standard & Aug. & Standard & Aug.\\
\hline
Natural & 99.31\% & 99.53\% & 92.62\% & 90.02\% & 75.96\% & 65.96\% \\
Worst-of-10 & 73.32\% & 98.33\% & 20.13\% & 79.92\% & 47.83\% & 50.62\% \\
First-Order & 79.84\% & 98.78\% & 62.69\% & 85.92\% & 63.12\% & 66.05\% \\
Grid & \textbf{26.02}\% & \textbf{95.79}\% & \textbf{2.80}\% & \textbf{58.92}\% & \textbf{31.42}\% & \textbf{32.90}\%
\end{tabular}}
\end{center}
\end{table*}

\paragraph{Understanding the Failure of First-Order Methods.}
The fact that first-order methods fail to reliably find adversarial rotations
and translations is in sharp contrast to previous work on
$\ell_p$-robustness~\citep{carlini2017towards, madry2018towards}.
For $\ell_p$-bounded perturbations parametrized directly in pixel space, prior work found the optimization landscape to be well-behaved which allowed first-order methods to consistently find maxima with high loss.
In the case of spatial perturbations, we observe that the non-concavity of the problem is a significant barrier for first-order methods.
We investigate this issue by visualizing the loss landscape.
For a few random examples from the three datasets, we plot the cross-entropy loss of the examples as a function of translation and rotation.
Figure~\ref{fig:landscape} shows one example for each dataset and additional examples are visualized in Figure~\ref{fig:landscape_complete} of the appendix.
The plots show that the loss landscape is indeed non-concave and contains many local maxima of low value.
The low-dimensional problem structure seems to make non-concavity a crucial obstacle.
Even for MNIST, where we observe fewer local maxima, the large flat regions prevent first-order methods from finding transformations of high loss.

\paragraph{Relation to Black-Box Attacks.}
Given its limited interaction with the model, the worst-of-10 adversary achieves a significant reduction in classification accuracy.
It performs only 10 \emph{random}, \emph{non-adaptive} queries to the model and is still able to find adversarial examples for a large fraction of the inputs (see Table \ref{tab:methods_small}).
The low query complexity is an important baseline for black-box attacks on neural networks, which recently gained significant interest~\citep{papernot2017practical,chen2017zoo,bhagoji2017exploring,ilyas2018blackbox}.
Black-box attacks rely only function evaluations of the target classifier, without additional information such as gradients.
The main challenge is to construct an adversarial example from a small number of queries.
Our results show that it is possible to find adversarial rotations and translations for a significant fraction of inputs with very few queries.

\paragraph{Combining Spatial and $\ell_\infty$-Bounded Perturbations}
Table~\ref{tab:core} shows that models trained to be robust to
$\ell_\infty$-bounded perturbations do not achieve higher robustness to spatial perturbations.
This provides evidence that the two families of perturbation are orthogonal to each other.
We further investigate this possibility by considering a combined adversary that utilizes $\ell_\infty$-bounded perturbations on top of rotations and translations.
The results are shown in Figure~\ref{fig:linf}.
We indeed observe that these combined attacks reduce classification accuracy in an (approximately) additive manner.

\subsection{Evaluating Our Defense Methods.}
As we see in Table~\ref{tab:core}, training with a worst-of-10 adversary significantly increases the spatial robustness of the model, also compared to data augmentation with random transformations.
We conjecture that using more reliable methods to compute the worst-case transformations will further improve these results.
Unfortunately, increasing the number of random transformations per training example quickly becomes computationally expensive.
And as pointed out above, current first-order methods also appear to be insufficient for finding worst-case transformations efficiently.

Our results for majority-based inference are presented in Table~\ref{tab:majority} of Appendix~\ref{app:omitted}.
By combining these two defenses, we improve the worst-case performance of the models from 26\% to 98\% on MNIST, from 3\% to 82\% on CIFAR10, and from 31\% to 56\% on ImageNet (Top~1).

\section{Conclusions}
We examined the robustness of state-of-the-art image classifiers to translations and rotations. 
We observed that even a small number of randomly chosen perturbations of the input are sufficient to considerably degrade the classifier's performance.

The fact that common neural networks are vulnerable to simple and naturally occurring spatial transformations (and that these transformations can be found easily from just a few random tries) indicates that adversarial robustness should be a concern not only in a fully worst-case security setting.
We conjecture that additional techniques need to be incorporated in the architecture and training procedures of modern classifiers to achieve worst-case spatial robustness.
Also, our results underline the need to consider broader notions of similarity than only pixel-wise distances when studying adversarial misclassification attacks.
In particular, we view combining the pixel-wise distances with rotations and translations as a next step towards the ``right'' notion of similarity in the context of images.

\section*{Acknowledgements}
Dimitris Tsipras was supported in part by the NSF grant CCF-1553428 and the NSF
Frontier grant CNS-1413920. Aleksander M\k{a}dry was supported in part by an Alfred P. Sloan Research Fellowship, a Google Research Award, and the NSF grants CCF-1553428 and CNS-1815221.


\bibliography{bibliography/bib.bib}
\bibliographystyle{icml2019}
\clearpage
\appendix
\onecolumn
\section{Omitted Tables and Figures}
\label{app:omitted}
\begin{figure}[!htp]
\begin{center}
\includegraphics[width=.98\textwidth]{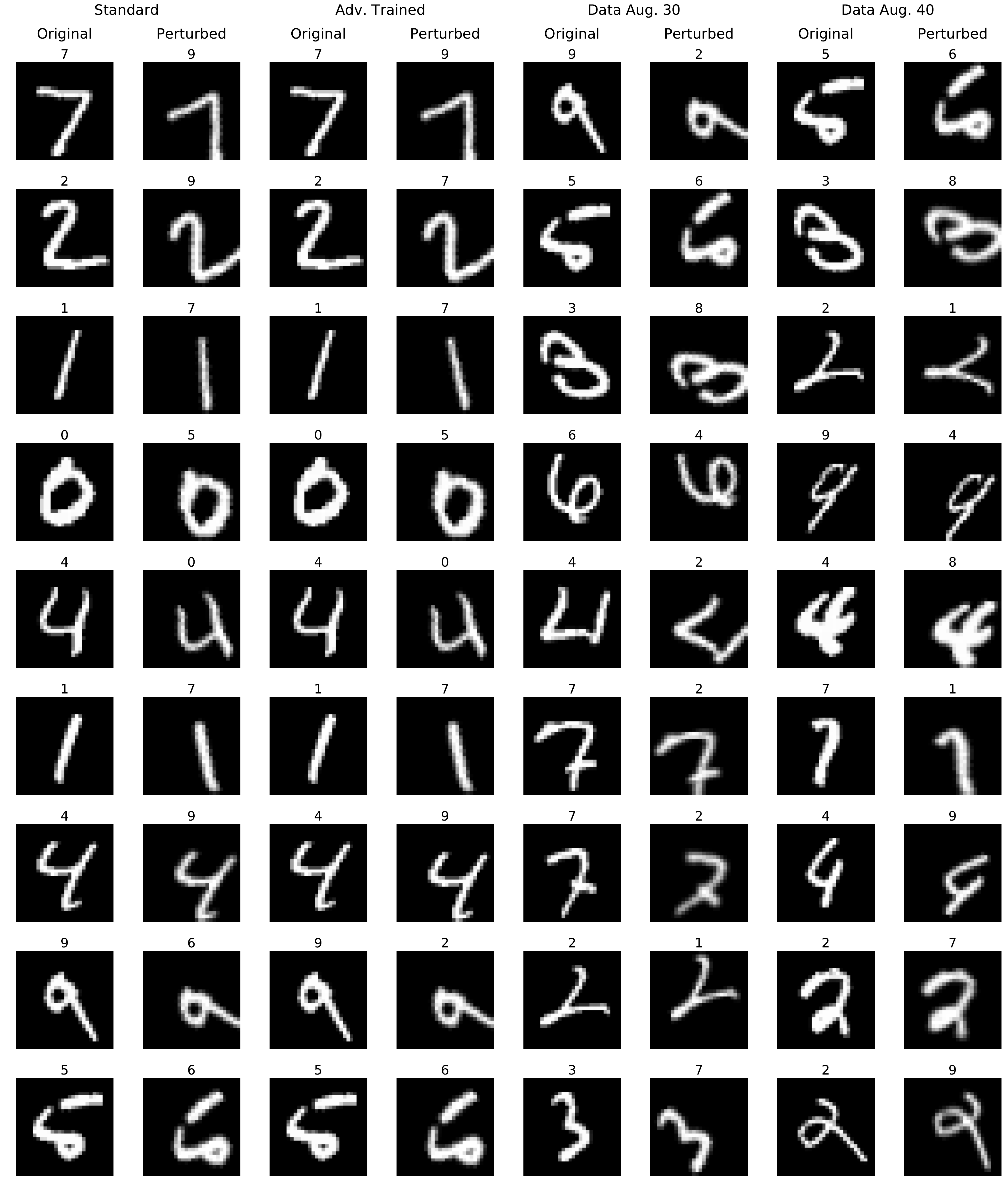}
\caption{MNIST. Successful adversarial examples for the models studied in Section~\ref{sec:exp}.
Rotations are restricted to be within $30\degree$ of the original image and translations up to $3$ pixels per direction (image size $28\times28$). Each example is visualized along with its predicted label in the original and perturbed versions.}
\label{fig:mnist_fool}
\end{center}
\end{figure}

\begin{figure}[!htp]
\begin{center}
\includegraphics[width=\textwidth]{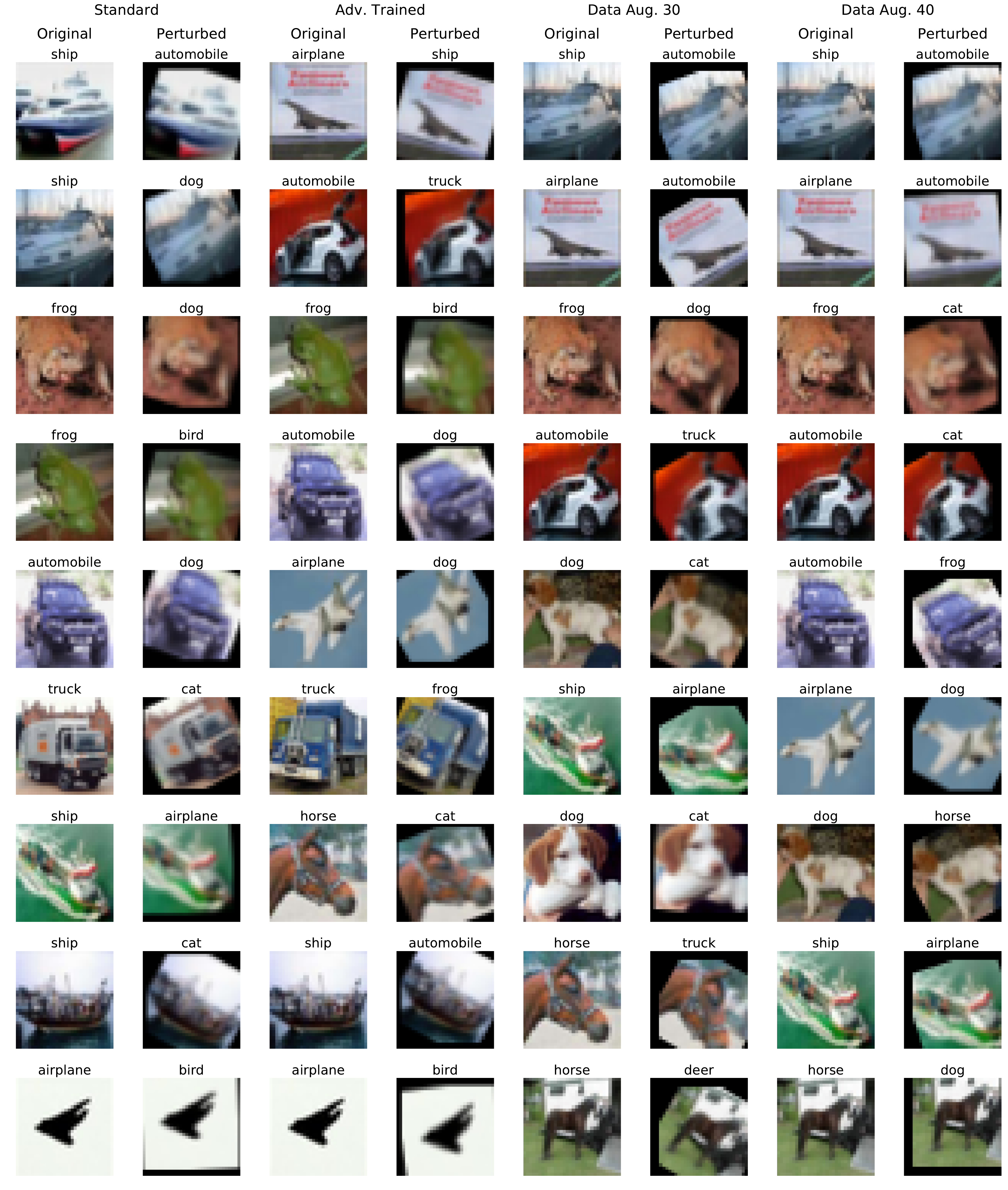}
\caption{CIFAR10. Successful adversarial examples for the models studied in Section~\ref{sec:exp}.
Rotations are restricted to be within $30\degree$ of the original and translations up to $3$ pixels per directions (image size $32\times32$). Each example is visualized along with its predicted label in the original and perturbed version.}
\label{fig:cifar_fool}
\end{center}
\end{figure}

\begin{figure}[!htp]
\begin{center}
\includegraphics[width=\textwidth]{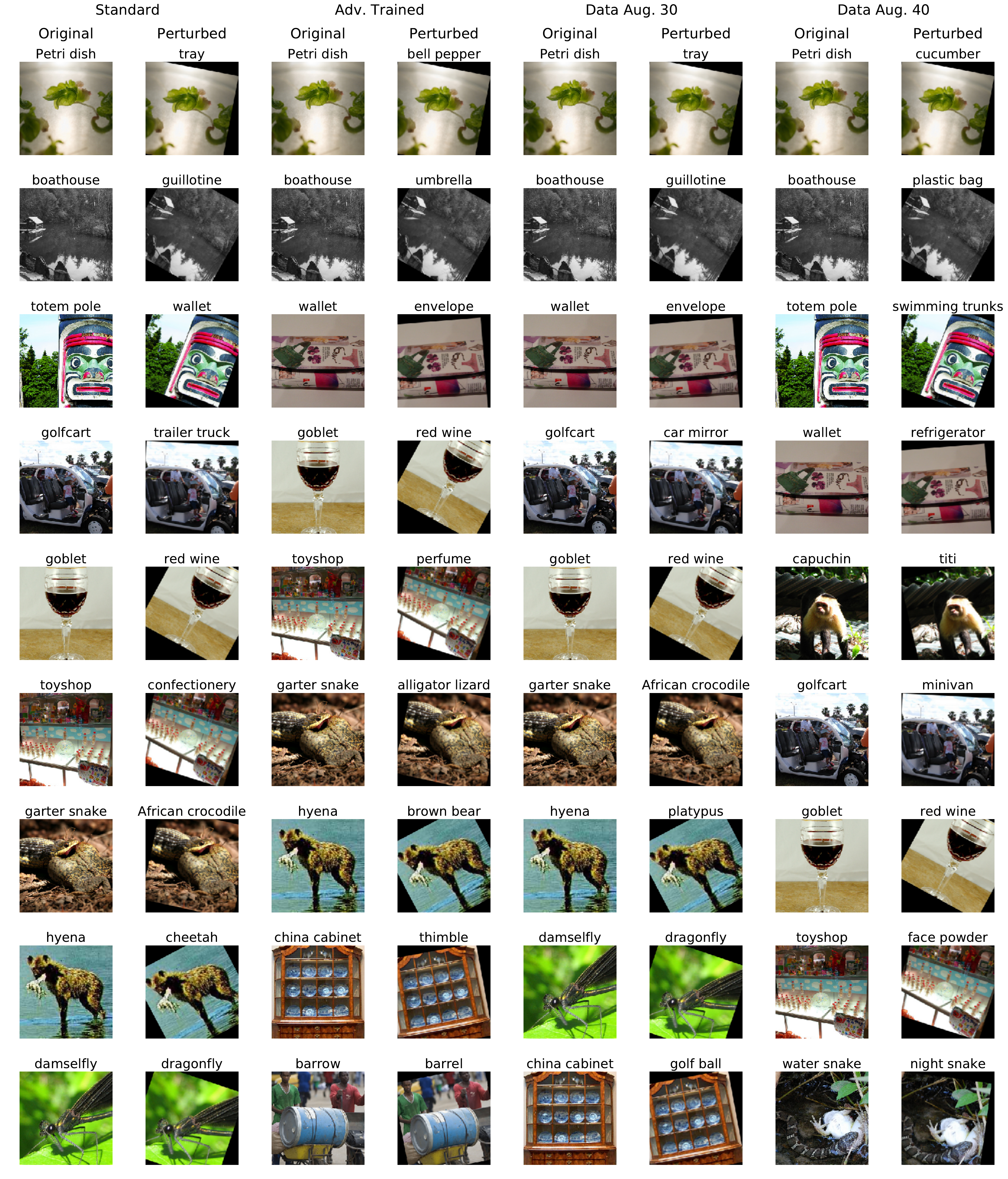}
\caption{ImageNet. Successful adversarial examples for the models studied in Section~\ref{sec:exp}.
Rotations are restricted to be within $30\degree$ of the original and translations up to $24$ pixels per directions (image size $299\times299$). Each example is visualized along with its predicted label in the original and perturbed version.}
\label{fig:imagenet_fool}
\end{center}
\end{figure}

\begin{figure}[!htp]
\begin{center}
\includegraphics[width=0.70\textwidth]{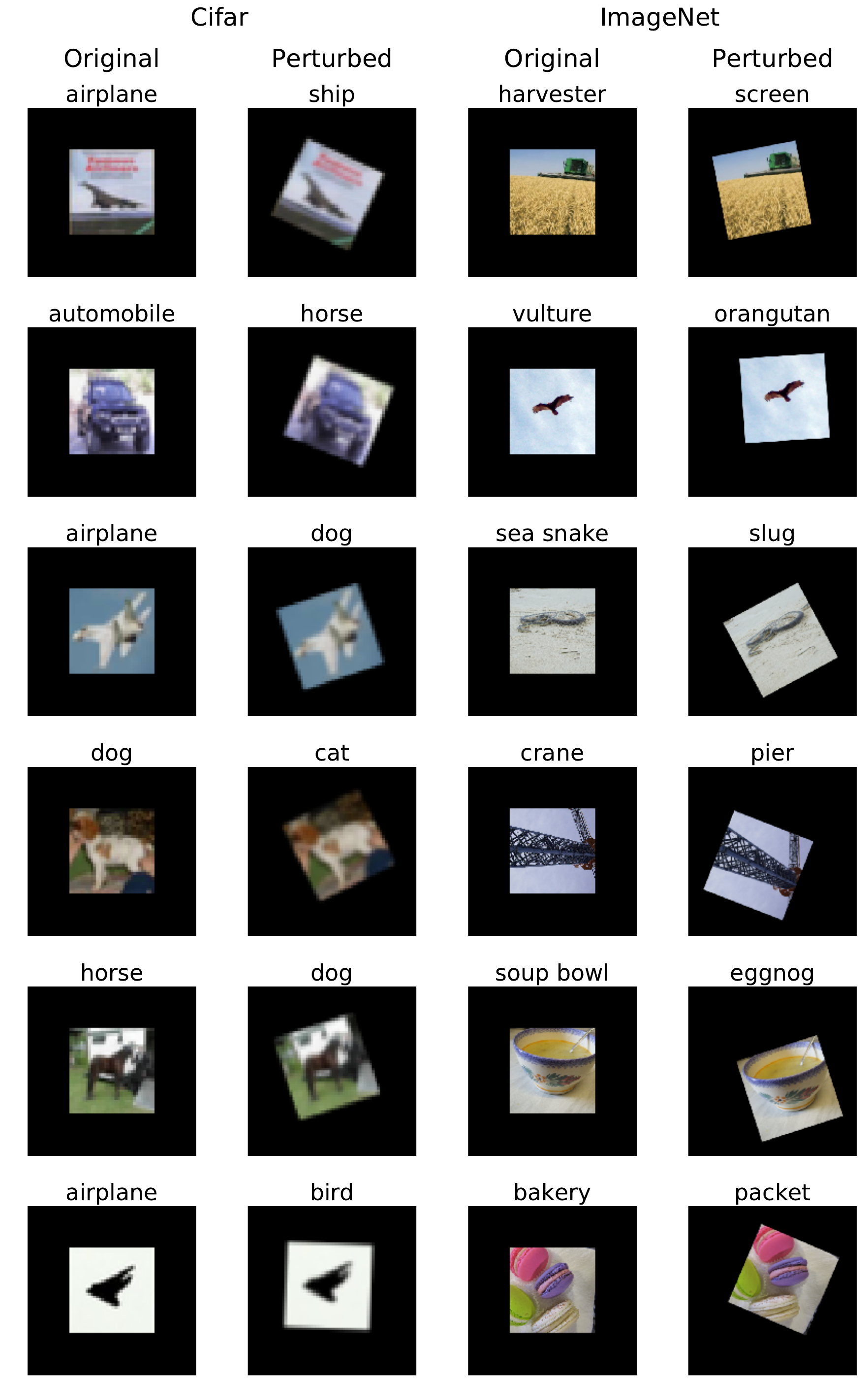}
\caption{Sample adversarial transformations for the "black-canvas" setting for the standard models on CIFAR10 and ImageNet.}
\label{fig:canvas_fool}
\end{center}
\end{figure}

\begin{figure}[!htp]
\begin{center}
\includegraphics[width=0.30\textwidth]{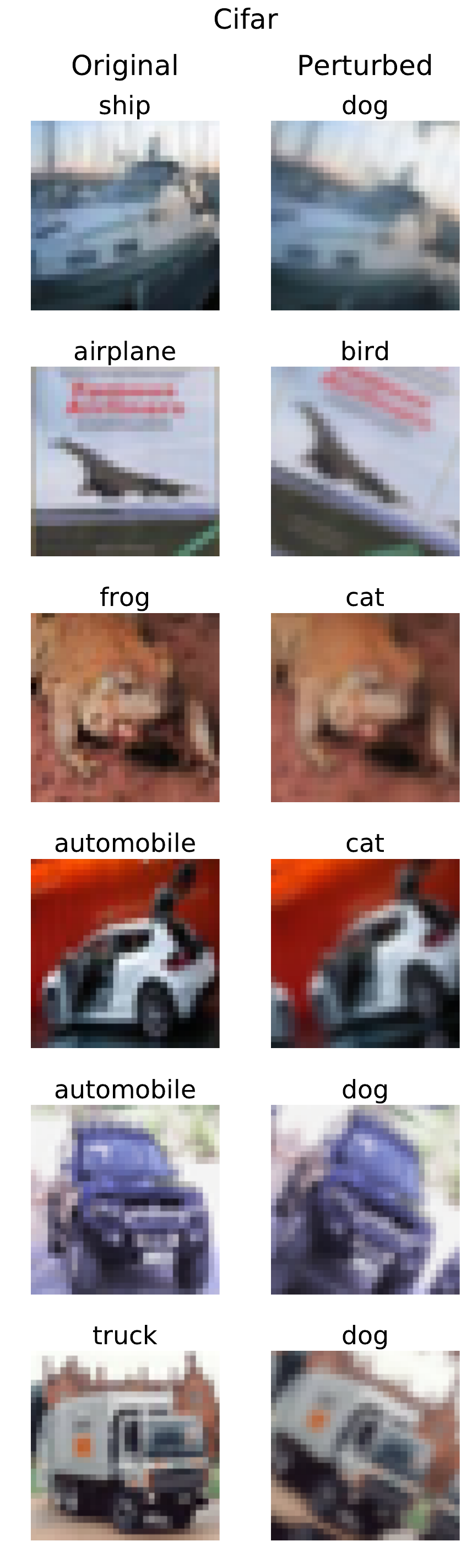}
\caption{Sample adversarial transformations for the reflection padding setting for the standard models on CIFAR10.}
\label{fig:reflect_fool}
\end{center}
\end{figure}


\begin{figure}[!htp]
\begin{center}
MNIST\\
\includegraphics[scale=.3]{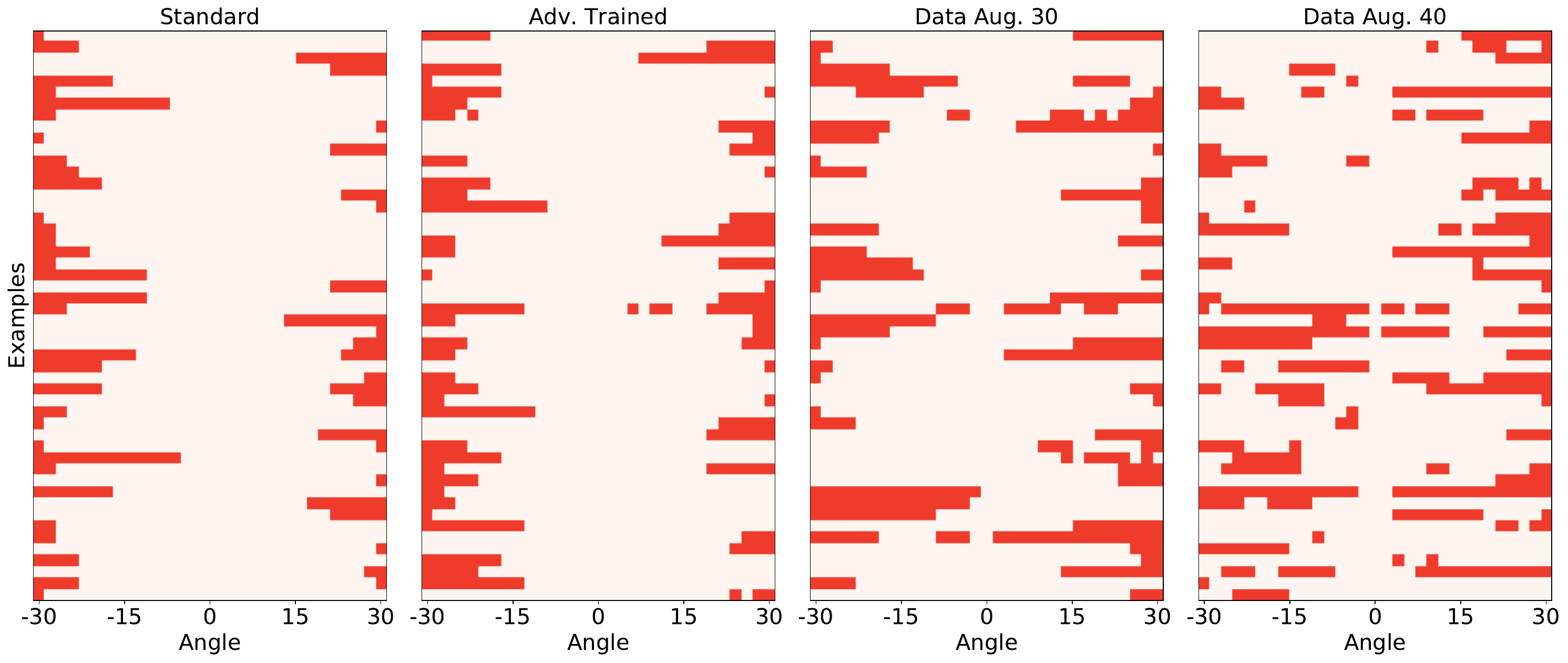}\\
CIFAR10\\
\includegraphics[scale=.3]{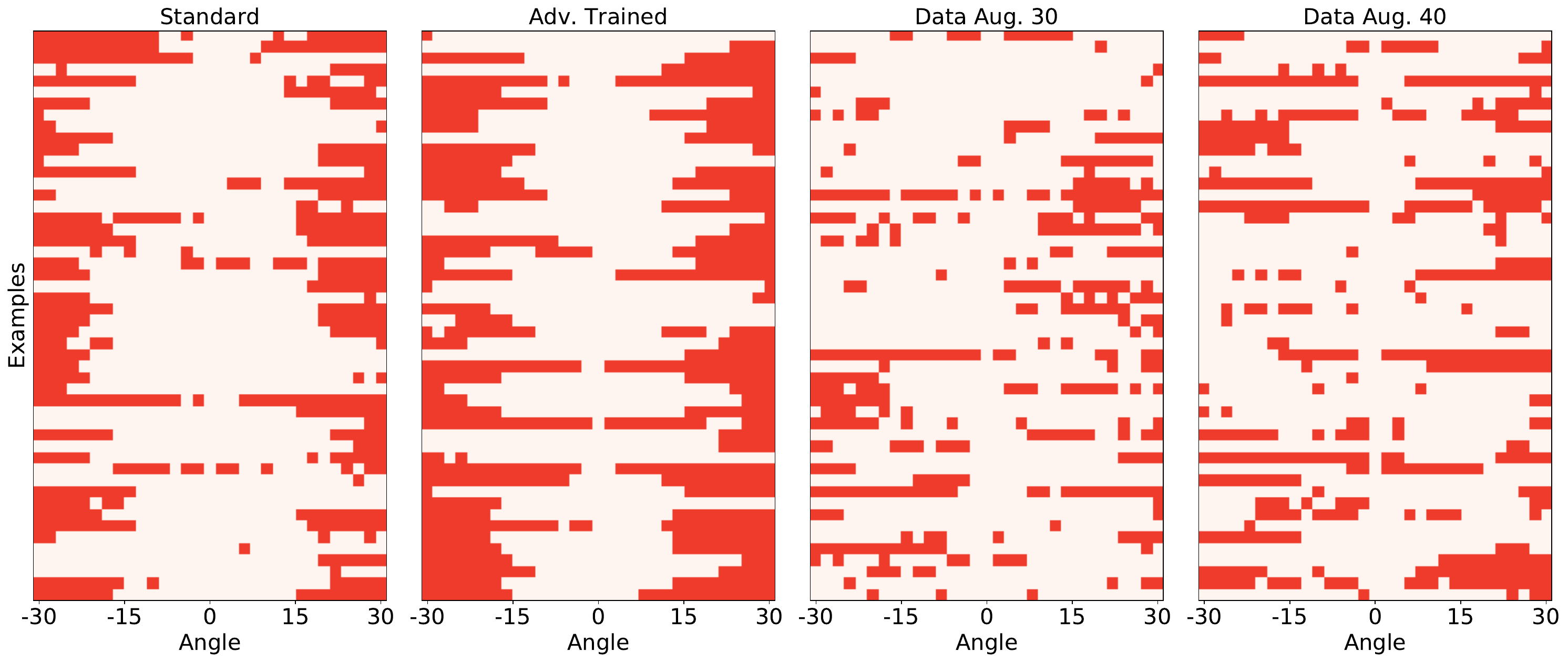}
\caption{Visualizing which angles fool the classifier for 50 random examples on
  CIFAR and MNIST.
  For each dataset and model, we visualize one example per row.
  \textcolor{red}{Red} corresponds to \emph{misclassification} of the images. We
  observe that the angles fooling the models form a highly non-convex set.}
\label{fig:angles}
\end{center}
\end{figure}

\begin{figure}[!htp]
\begin{center}
MNIST\\
\vspace{3pt}
\includegraphics[scale=.3]{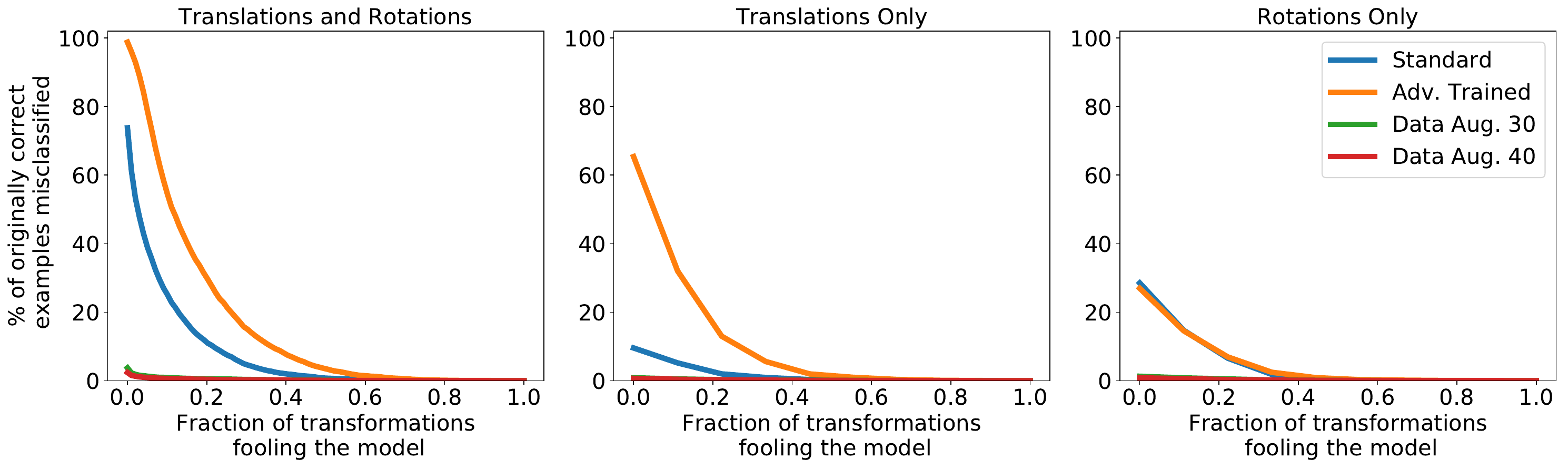}\\
\vspace{5pt}
CIFAR10\\
\vspace{3pt}
\includegraphics[scale=.3]{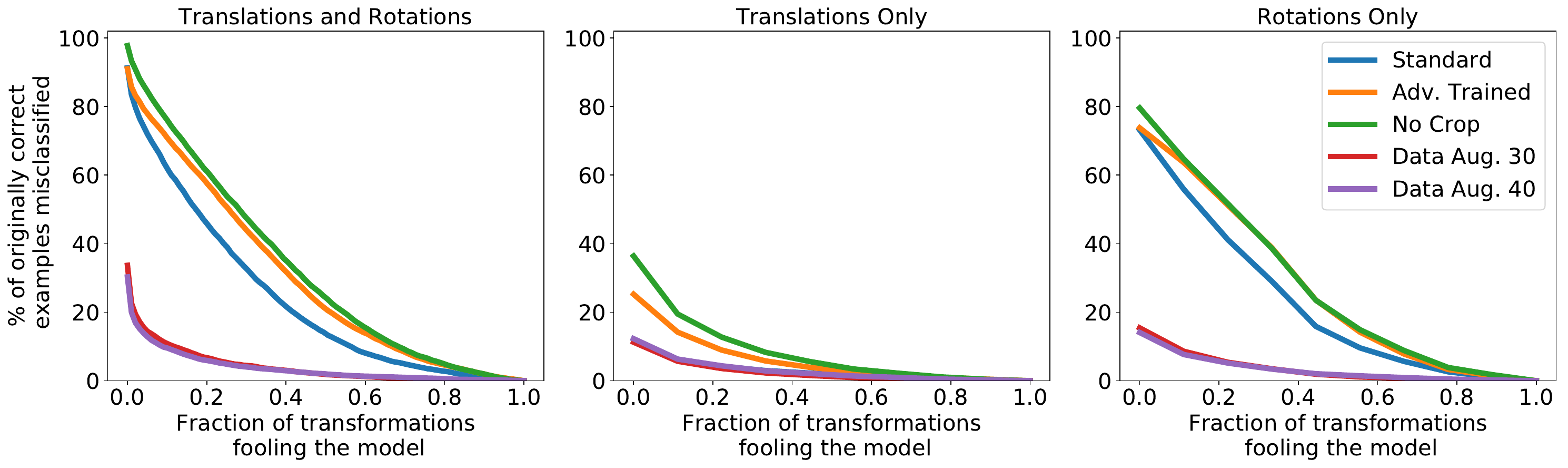}\\
\vspace{5pt}
ImageNet\\
\vspace{3pt}
\includegraphics[scale=.3]{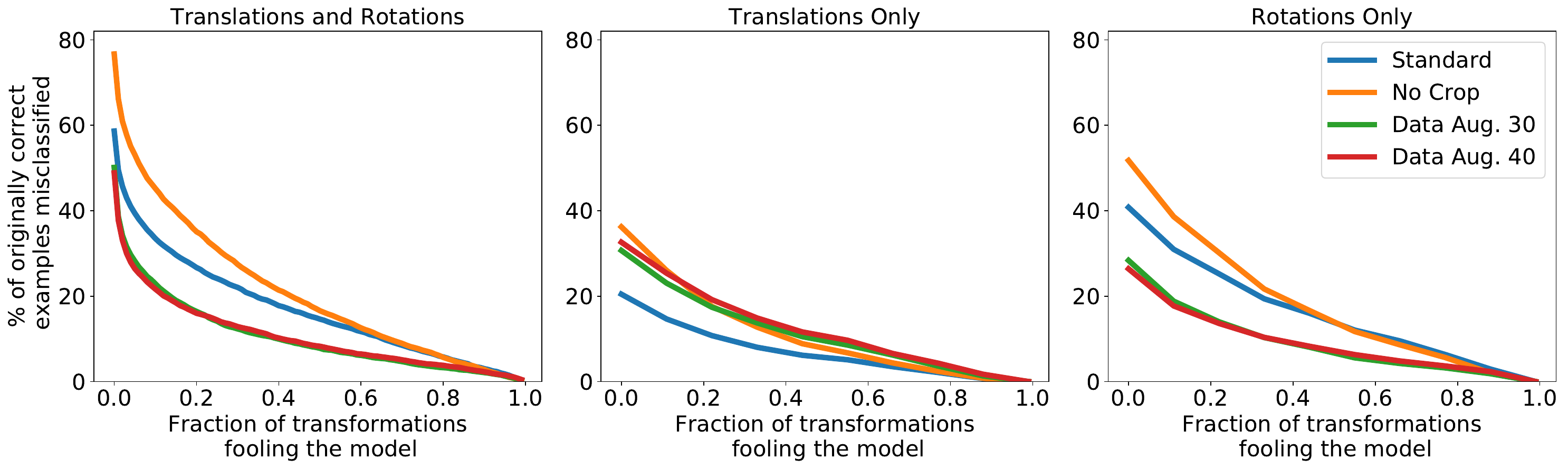}
\caption{Cumulative Density Function plots. For each fraction of grid points $p$, we plot the percentage of correctly classified test set examples that are fooled by at least $p$ of the grid points. For instance, we can see from the first plot, MNIST Translations and Rotations, that approximately $10$\% of the correctly classified natural examples are misclassified under $1/5$ of the grid points transformations.}
\label{fig:hist}
\end{center}
\end{figure}

\begin{table*}[htp]
\caption{
Comparison of attack methods across datasets and models.
}
\label{tab:methods}
\begin{center}
{\renewcommand{\arraystretch}{1.1}
\begin{tabular}{c|c|cccc}
    & Model & Natural & Worst-of-10 & FO & Grid \\
\hline                                                                            
\multirow{4}{*}{\rotatebox[origin=c]{90}{MNIST}}
& Standard
    & 99.31\% & 73.32\% & 79.84\% & \textbf{26.02}\%\\
& $\ell_\infty$-Adversarially Trained
    & 98.65\% & 51.18\%  & 81.23\% & \textbf{1.20}\%\\
& Aug. 30 ($\pm 3\px,\pm30\degree$)
    & 99.53\% & 98.33\%  & 98.78\% & \textbf{95.79}\%\\
& Aug. 40 ($\pm 4\px,\pm40\degree$)
    & 99.34\% & 98.49\%  & 98.74\% & \textbf{96.95}\%\\

\hline
\multirow{5}{*}{\rotatebox[origin=c]{90}{CIFAR10}}
& Standard
    & 92.62\% & 20.13\%  & 62.69\% & \textbf{ 2.80}\%\\
& No Crop
    & 90.34\% & 15.04\%   & 52.27\% & \textbf{ 1.86}\%\\
& $\ell_\infty$-Adversarially Trained
    & 80.21\% & 19.38\%  & 33.24\% & \textbf{ 6.02}\%\\
& Aug. 30 $(\pm3\px,\pm30\degree)$
    & 90.02\% & 79.92\%  & 85.92\% & \textbf{58.92}\%\\
& Aug. 40 $(\pm4\px,\pm40\degree)$
    & 88.83\% & 80.47\%  & 85.48\% & \textbf{61.69}\%\\

\hline                                                                              
\multirow{4}{*}{\rotatebox[origin=c]{90}{ImageNet}}
& Standard                          
    & 75.96\%  & 47.83\%   & 63.12\% & \textbf{31.42}\%\\
& No Crop                           
    & 70.81\%  & 35.52\%   & 55.93\% & \textbf{16.52}\%\\
& Aug. 30 $(\pm 24\px,\pm 30\degree)$
    & 65.96\%  & 50.62\%   & 66.05\% & \textbf{32.90}\%\\
& Aug. 40 $(\pm 32\px,\pm 40\degree)$
    & 66.19\%  & 51.11\%   & 66.14\% & \textbf{33.86}\%\\
\end{tabular}}
\end{center}
\end{table*}

\begin{figure}[!htp]
\begin{center}
\includegraphics[width=0.95\textwidth]{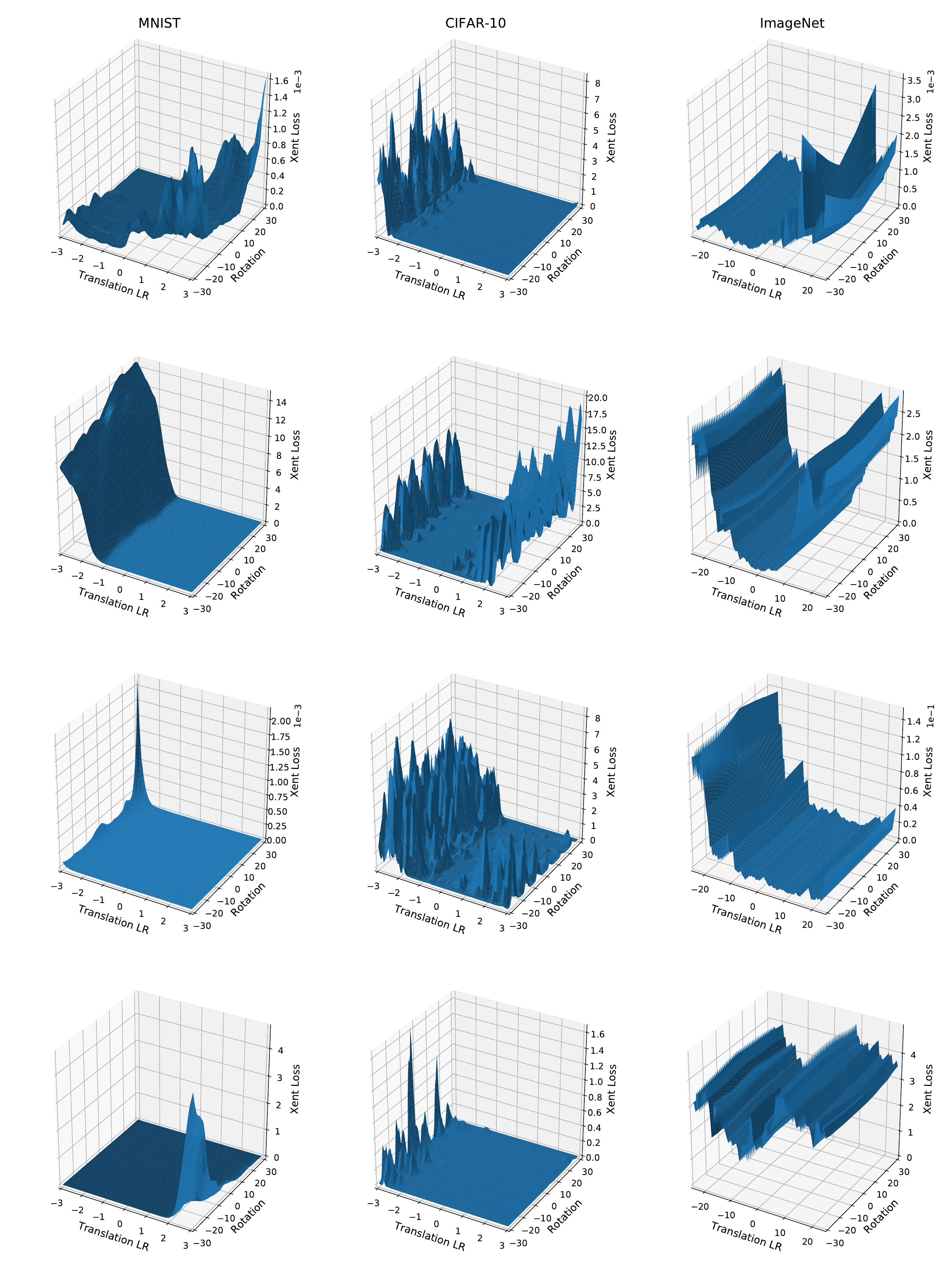}
\caption{Loss landscape of $4$ random examples for each dataset when performing left-right translations and rotations.
Translations and rotations are restricted to $10$\% of the image pixels and $30\degree$ respectively.
We observe that the landscape is significantly non-concave, making rendering FO methods for adversarial example generation powerless.}
\label{fig:landscape_complete}
\end{center}
\end{figure}

\begin{table*}[htp]
\caption{
Evaluation of a subset of Table~\ref{tab:core} in the ``black-canvas'' setting (images are zero-padded to avoid cropping due to rotations and translations). The models are trained on padded images.
}
\label{tab:canvas}
\begin{center}
{\renewcommand{\arraystretch}{1.1}
\begin{tabular}{c|c|ccccccc}
    & & Natural & Random & Worst-of-10 & Grid & Trans. Grid & Rot. Grid \\
\hline                                                                            
\multirow{4}{*}{\rotatebox[origin=c]{90}{CIFAR10}}
& Standard
    & 91.81\% & 70.23\% & 25.51\% & \textbf{ 6.55}\% & 83.38\% & 12.44\% \\
& No Crop
    & 89.70\% & 52.86\% & 14.14\% & \textbf{ 1.17}\% & 47.94\% &  9.46\% \\
& Aug. 30 $(\pm3\px,\pm30\degree)$
    & 91.45\% & 90.82\% & 80.53\% & \textbf{63.64}\% & 82.28\% & 76.32\% \\
& Aug. 40 $(\pm4\px,\pm40\degree)$
    & 91.24\% & 91.00\% & 81.81\% & \textbf{66.64}\% & 81.75\% & 78.57\%  \\
\hline                                                                     
\multirow{4}{*}{\rotatebox[origin=c]{90}{ImageNet}}
& Standard
    & 73.60\% & 46.59\% & 29.51\% & \textbf{15.38}\% & 28.03\% & 23.81\% \\
& No Crop
    & 66.28\% & 38.70\% & 14.17\% & \textbf{ 3.43}\% &  8.87\% & 10.97\% \\
& Aug. 30 $(\pm 24\px,\pm 30\degree)$
    & 64.60\% & 67.75\% & 47.32\% & \textbf{28.51}\% & 45.33\% & 39.33\% \\
& Aug. 40 $(\pm 32\px,\pm 40\degree)$
    & 49.20\% & 57.69\% & 38.36\% & \textbf{22.10}\% & 32.84\% & 32.95\% \\
\end{tabular}}
\end{center}
\end{table*}

\clearpage
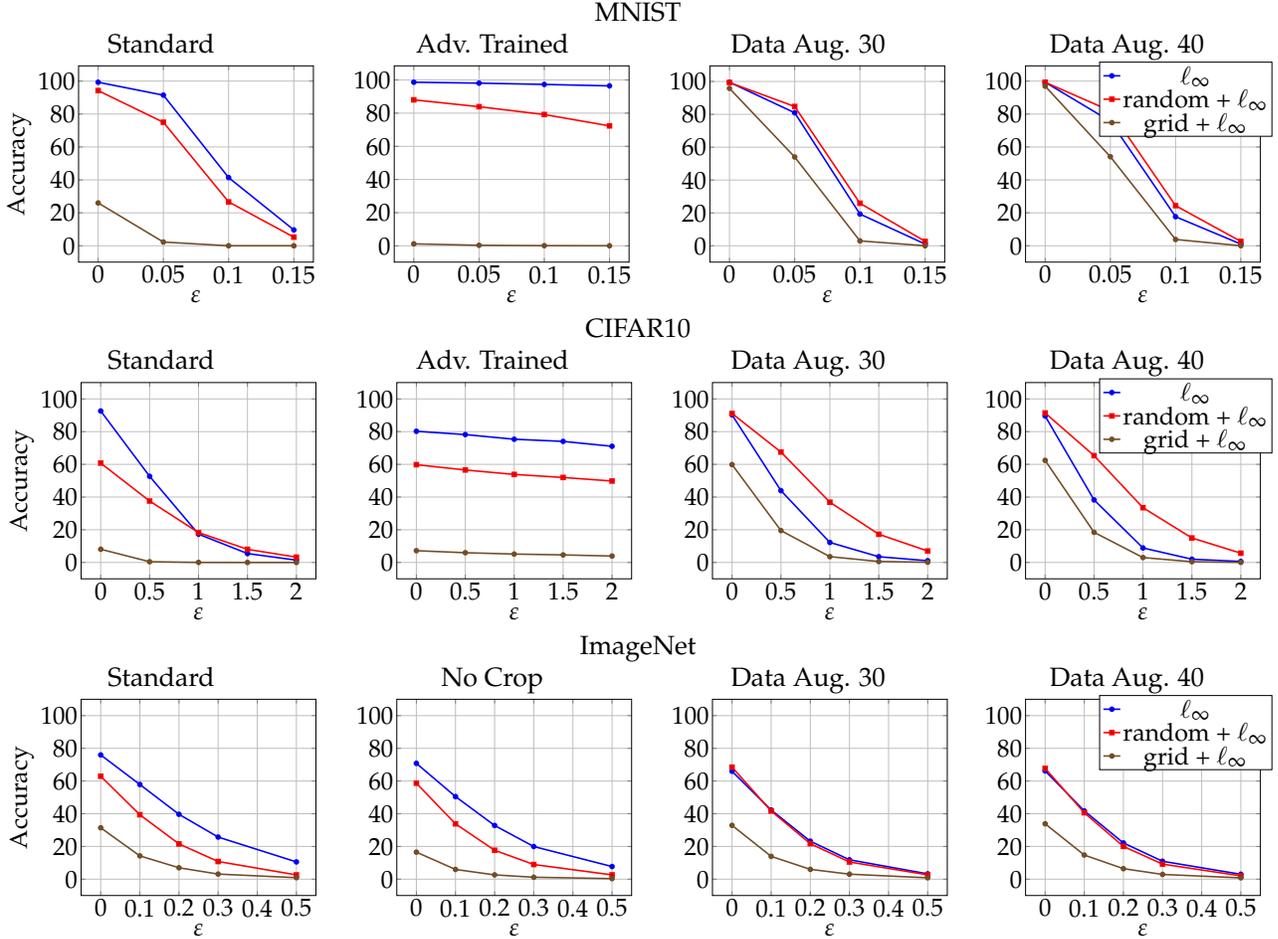
\begin{figure*}[htp]
\begin{center}
\begin{tabular}{cccc}
\multicolumn{4}{c}{ MNIST }\\
Standard & Adv. Trained & Data Aug. 30 & Data Aug. 40 \\
\begin{tikzpicture}[scale=0.375]
    \begin{axis}[
        xlabel=$\eps$,
        ylabel=Accuracy,
        ylabel near ticks,
        xtick = data,
        grid = both,
        legend pos=north east,
        label style={font=\Huge},
        x label style={at={(axis description cs:0.5,-0.05)},anchor=north},
        tick label style={font=\Huge, /pgf/number format/fixed},
        every axis plot/.append style={ultra thick},
        width=0.6\textwidth
    ]
    \addplot plot coordinates {
        (0.15,    9.62)
        (0.1,    41.41)
        (0.05,   91.46)
        (0,      99.31)
    };
    \addplot plot coordinates {
        (0.15,    5.24)
        (0.1,    26.64)
        (0.05,   75.02)
        (0,      94.23)
    };
    \addplot plot coordinates {
        (0.15,    0.00)
        (0.1,     0.00)
        (0.05,    2.26)
        (0,      26.02)
    };

    \end{axis}
\end{tikzpicture} &
\begin{tikzpicture}[scale=0.375]
    \begin{axis}[
        xlabel=$\eps$,
        ylabel near ticks,
        xtick = data,
        grid = both,
        legend pos=north east,
        label style={font=\Huge},
        x label style={at={(axis description cs:0.5,-0.05)},anchor=north},
        tick label style={font=\Huge, /pgf/number format/fixed},
        every axis plot/.append style={ultra thick},
        width=0.6\textwidth
    ]
    \addplot plot coordinates {
        (0.15,   96.45)
        (0.1,    97.32)
        (0.05,   98.10)
        (0,      98.65)
    };
    \addplot plot coordinates {
        (0.15,   72.33)
        (0.1,    79.18)
        (0.05,   83.89)
        (0,      88.02)
    };
    \addplot plot coordinates {
        (0.15,   0.05)
        (0.1,    0.11)
        (0.05,   0.33)
        (0,      1.20)
    };

    \end{axis}
\end{tikzpicture} &
\begin{tikzpicture}[scale=.375]
    \begin{axis}[
        xlabel=$\eps$,
        ylabel near ticks,
        xtick = data,
        grid = both,
        legend pos=north east,
        label style={font=\Huge},
        x label style={at={(axis description cs:0.5,-0.05)},anchor=north},
        tick label style={font=\Huge, /pgf/number format/fixed},
        every axis plot/.append style={ultra thick},
        width=0.6\textwidth
    ]
    \addplot plot coordinates {
        (0.15,    1.05)
        (0.1,    19.30)
        (0.05,   80.95)
        (0,      99.53)
    };
    \addplot plot coordinates {
        (0.15,    2.66)
        (0.1,    25.87)
        (0.05,   84.76)
        (0,      99.35)
    };
    \addplot plot coordinates {
        (0.15,    0.00)
        (0.1,     2.96)
        (0.05,   54.02)
        (0,      95.79)
    };

    \end{axis}
\end{tikzpicture} &
\begin{tikzpicture}[scale=.375]
    \begin{axis}[
        xlabel=$\eps$,
        ylabel near ticks,
        xtick = data,
        grid = both,
        legend pos=north east,
        label style={font=\Huge},
        x label style={at={(axis description cs:0.5,-0.05)},anchor=north},
        tick label style={font=\Huge, /pgf/number format/fixed},
        every axis plot/.append style={ultra thick},
        width=0.6\textwidth,
        legend style = {at={(1.05,1.02)}, font=\Huge}
    ]
    \addplot plot coordinates {
        (0.15,    1.05)
        (0.1,    17.65)
        (0.05,   75.66)
        (0,      99.34)
    };
    \addplot plot coordinates {
        (0.15,    2.65)
        (0.1,    24.36)
        (0.05,   81.96)
        (0,      99.31)
    };
    \addplot plot coordinates {
        (0.15,    0.00)
        (0.1,     3.82)
        (0.05,   54.10)
        (0,      96.95)
    };
    \legend{$\ell_\infty$, random + $\ell_\infty$, grid + $\ell_\infty$}

    \end{axis}
\end{tikzpicture} \\
\multicolumn{4}{c}{ CIFAR10 }\\
Standard & Adv. Trained & Data Aug. 30 & Data Aug. 40 \\
\begin{tikzpicture}[scale=0.375]
    \begin{axis}[
        xlabel=$\eps$,
        ylabel=Accuracy,
        ylabel near ticks,
        ymin=-10,ymax=110,
        xtick = data,
        grid = both,
        legend pos=north east,
        label style={font=\Huge},
        x label style={at={(axis description cs:0.5,-0.05)},anchor=north},
        tick label style={font=\Huge, /pgf/number format/fixed},
        every axis plot/.append style={ultra thick},
        width=0.6\textwidth
    ]
    \addplot plot coordinates {
        (  0,   92.62)
        (0.5,   52.68)
        (  1,   17.30)
        (1.5,    5.45)
        (  2,    1.42)
    };
    \addplot plot coordinates {
        (  0,   60.76)
        (0.5,   37.58)
        (  1,   18.26)
        (1.5,    8.03)
        (  2,    3.25)
    };
    \addplot plot coordinates {
        (  0,    8.08)
        (0.5,    0.45)
        (  1,    0.03)
        (1.5,    0.00)
        (  2,    0.00)
    };
    \end{axis}
\end{tikzpicture} &
\begin{tikzpicture}[scale=0.375]
    \begin{axis}[
        xlabel=$\eps$,
        ylabel near ticks,
        ymin=-10,ymax=110,
        xtick = data,
        grid = both,
        legend pos=north east,
        label style={font=\Huge},
        x label style={at={(axis description cs:0.5,-0.05)},anchor=north},
        tick label style={font=\Huge, /pgf/number format/fixed},
        every axis plot/.append style={ultra thick},
        width=0.6\textwidth
    ]
    \addplot plot coordinates {
        (  0,    80.21)
        (0.5,    78.20)
        (  1,    75.35)
        (1.5,    74.03)
        (  2,    71.04)
    };
    \addplot plot coordinates {
        (  0,    59.79)
        (0.5,    56.55)
        (  1,    53.83)
        (1.5,    51.98)
        (  2,    49.79)
    };
    \addplot plot coordinates {
        (  0,    7.20)
        (0.5,    5.96)
        (  1,    5.14)
        (1.5,    4.62)
        (  2,    3.92)
    };
    \end{axis}
\end{tikzpicture} &
\begin{tikzpicture}[scale=.375]
    \begin{axis}[
        xlabel=$\eps$,
        ylabel near ticks,
        ymin=-10,ymax=110,
        xtick = data,
        grid = both,
        legend pos=north east,
        label style={font=\Huge},
        x label style={at={(axis description cs:0.5,-0.05)},anchor=north},
        tick label style={font=\Huge, /pgf/number format/fixed},
        every axis plot/.append style={ultra thick},
        width=0.6\textwidth
    ]
    \addplot plot coordinates {
        (  0,    90.25)
        (0.5,    43.92)
        (  1,    12.23)
        (1.5,     3.51)
        (  2,     1.06)
    };
    \addplot plot coordinates {
        (  0,    91.09)
        (0.5,    67.52)
        (  1,    36.85)
        (1.5,    17.24)
        (  2,     7.00)
    };
    \addplot plot coordinates {
        (  0,    59.87)
        (0.5,    19.48)
        (  1,     3.55)
        (1.5,     0.59)
        (  2,     0.14)
    };
    \end{axis}
\end{tikzpicture} &
\begin{tikzpicture}[scale=.375]
    \begin{axis}[
        xlabel=$\eps$,
        ylabel near ticks,
        ymin=-10,ymax=110,
        xtick = data,
        grid = both,
        legend pos=north east,
        label style={font=\Huge},
        x label style={at={(axis description cs:0.5,-0.05)},anchor=north},
        tick label style={font=\Huge, /pgf/number format/fixed},
        every axis plot/.append style={ultra thick},
        width=0.6\textwidth,
        legend style = {at={(1.05,1.02)}, font=\Huge}
    ]
    \addplot plot coordinates {
        (  0,    89.55)
        (0.5,    38.26)
        (  1,     8.87)
        (1.5,     2.00)
        (  2,     0.62)
    };
    \addplot plot coordinates {
        (  0,    91.40)
        (0.5,    65.32)
        (  1,    33.50)
        (1.5,    14.97)
        (  2,     5.69)
    };
    \addplot plot coordinates {
        (  0,    62.42)
        (0.5,    18.44)
        (  1,     3.00)
        (1.5,     0.45)
        (  2,     0.06)
    };
    \legend{$\ell_\infty$, random + $\ell_\infty$, grid + $\ell_\infty$}
    \end{axis}
\end{tikzpicture} \\
\multicolumn{4}{c}{ImageNet}\\
Standard & No Crop & Data Aug. 30 & Data Aug. 40 \\
\begin{tikzpicture}[scale=0.375]
    \begin{axis}[
        xlabel=$\eps$,
        ylabel=Accuracy,
        ylabel near ticks,
        ymin=-10,ymax=110,
        grid = both,
        legend pos=north east,
        label style={font=\Huge},
        x label style={at={(axis description cs:0.5,-0.05)},anchor=north},
        tick label style={font=\Huge, /pgf/number format/fixed},
        every axis plot/.append style={ultra thick},
        width=0.6\textwidth
    ]
    \addplot plot coordinates {
        (0.0, 75.96)
        (0.1, 57.90)
        (0.2, 39.66)
        (0.3, 25.74)
        (0.5, 10.55)
    };
    \addplot plot coordinates {
        (0.0, 62.92)
        (0.1, 39.41)
        (0.2, 21.55)
        (0.3, 10.80)
        (0.5,  2.60)
    };
    \addplot plot coordinates {
        (0.0, 31.42)
        (0.1, 14.23)
        (0.2,  6.97)
        (0.3,  3.11)
        (0.5,  0.91)
    };
    \end{axis}
\end{tikzpicture} &
\begin{tikzpicture}[scale=0.375]
    \begin{axis}[
        xlabel=$\eps$,
        ylabel near ticks,
        ymin=-10,ymax=110,
        grid = both,
        legend pos=north east,
        label style={font=\Huge},
        x label style={at={(axis description cs:0.5,-0.05)},anchor=north},
        tick label style={font=\Huge, /pgf/number format/fixed},
        every axis plot/.append style={ultra thick},
        width=0.6\textwidth
    ]
    \addplot plot coordinates {
        (0.0, 70.81)
        (0.1, 50.50)
        (0.2, 32.82)
        (0.3, 19.95)
        (0.5,  7.72)
    };
    \addplot plot coordinates {
        (0.0, 58.66)
        (0.1, 33.82)
        (0.2, 17.66)
        (0.3,  8.96)
        (0.5,  2.62)
    };
    \addplot plot coordinates {
        (0.0, 16.52)
        (0.1,  5.94)
        (0.2,  2.62)
        (0.3,  1.17)
        (0.5,  0.33)
    };
    \end{axis}
\end{tikzpicture} &
\begin{tikzpicture}[scale=.375]
    \begin{axis}[
        xlabel=$\eps$,
        ylabel near ticks,
        ymin=-10,ymax=110,
        grid = both,
        legend pos=north east,
        label style={font=\Huge},
        x label style={at={(axis description cs:0.5,-0.05)},anchor=north},
        tick label style={font=\Huge, /pgf/number format/fixed},
        every axis plot/.append style={ultra thick},
        width=0.6\textwidth
    ]
    \addplot plot coordinates {
        (0.0, 65.96)
        (0.1, 42.28)
        (0.2, 23.23)
        (0.3, 11.85)
        (0.5,  3.24)
    };
    \addplot plot coordinates {
        (0.0, 68.49)
        (0.1, 41.68)
        (0.2, 21.71)
        (0.3, 10.49)
        (0.5,  2.65)
    };
    \addplot plot coordinates {
        (0.0, 32.90)
        (0.1, 13.94)
        (0.2,  6.00)
        (0.3,  3.06)
        (0.5,  0.83)
    };
    \end{axis}
\end{tikzpicture} &
\begin{tikzpicture}[scale=.375]
    \begin{axis}[
        xlabel=$\eps$,
        ylabel near ticks,
        ymin=-10,ymax=110,
        grid = both,
        legend pos=north east,
        label style={font=\Huge},
        x label style={at={(axis description cs:0.5,-0.05)},anchor=north},
        tick label style={font=\Huge, /pgf/number format/fixed},
        every axis plot/.append style={ultra thick},
        width=0.6\textwidth,
        legend style = {at={(1.05,1.02)}, font=\Huge}
    ]
    \addplot plot coordinates {
        (0.0, 66.19)
        (0.1, 41.76)
        (0.2, 22.15)
        (0.3, 10.94)
        (0.5,  2.99)
    };
    \addplot plot coordinates {
        (0.0, 67.72)
        (0.1, 40.61)
        (0.2, 20.04)
        (0.3,  9.20)
        (0.5,  2.03)
    };
    \addplot plot coordinates {
        (0.0, 33.86)
        (0.1, 14.75)
        (0.2,  6.42)
        (0.3,  2.90)
        (0.5,  0.80)
    };
    \legend{$\ell_\infty$, random + $\ell_\infty$, grid + $\ell_\infty$}
    \end{axis}
\end{tikzpicture} \\
\end{tabular}
\caption{Accuracy of different classifiers against $\ell_\infty$-bounded adversaries with various values of $\eps$ and spatial transformations.
For each value of $\eps$, we perform PGD to find the most adversarial $\ell_\infty$-bounded perturbation.
Additionally, we combine PGD with random rotations and translations and with a grid search over rotations and translations in order to find the transformation that combines with PGD in the most adversarial way. }
\label{fig:linf}
\end{center}
\end{figure*}

\begin{table}[!ht]
\caption{
Majority Defense.
Accuracy of different models on the natural evaluation set and against a combined rotation and translation adversary using aggregation of multiple random transformations. 
}
\label{tab:majority}
\begin{center}
{\renewcommand{\arraystretch}{1.1}
\setlength{\tabcolsep}{4.5pt}
\begin{tabular}{c|c|cc|cc}
& & \multicolumn{2}{c|}{Natural Acc.} & \multicolumn{2}{c}{Grid Acc.} \\
        & Model & Stand. & Vote         & Stand. & Vote \\
\hline
\multirow{5}{*}{\rotatebox[origin=c]{90}{MNIST}}
    & Standard  & 99.31\% & 98.71\% & 26.02\% & 18.80\% \\
    & Aug 30.    & \textbf{99.53}\% & 99.41\% & 95.79\% & 95.32\% \\
    & Aug 40.    & 99.34\% & 99.25\% & 96.95\% & 97.65\% \\
    & W-10 (30) & 99.48\% & 99.40\% & 97.32\% & 96.95\% \\
    & W-10 (40) & 99.42\% & 99.41\% & 97.88\% & \textbf{98.47}\% \\
\hline
\multirow{5}{*}{\rotatebox[origin=c]{90}{CIFAR10}}
    & Standard  & 92.62\% & 80.37\% &  2.82\% &  7.85\% \\
    & Aug 30.    & 90.02\% & 92.70\% & 58.90\% & 69.65\% \\
    & Aug 40.    & 88.83\% & 92.50\% & 61.69\% & 76.54\% \\
    & W-10 (30) & 91.34\% & 93.38\% & 69.17\% & 77.33\% \\
    & W-10 (40) & 91.00\% & \textbf{93.40}\% & 71.15\% & \textbf{81.52}\% \\
\hline
\multirow{5}{*}{\rotatebox[origin=c]{90}{ImageNet}}
    & Standard  & 75.96\% & 73.19\% & 31.42\% & 40.21\% \\
    & Aug 30.    & 65.96\% & 72.44\% & 32.90\% & 44.46\% \\
    & Aug 40.    & 66.19\% & 71.46\% & 33.86\% & 46.98\% \\
    & W-10 (30) & \textbf{76.14}\% & 74.92\% & 52.76\% & \textbf{56.45}\% \\
    & W-10 (40) & 74.64\% & 73.38\% & 50.23\% & 56.23\% \\
\end{tabular}
}
\end{center}
\end{table}

\section{Mirror Padding}
\label{app:reflection}
In the experiments of Section~\ref{sec:exp}, we filled the remaining pixels of rotated and translated images with black (also known as zero or constant padding).
This is the standard approach used when performing random cropping for data augmentation purposes.
We briefly examined the effect of mirror padding, that is replacing empty pixels by reflecting the image around the border\footnote{\url{https://www.tensorflow.org/api_docs/python/tf/pad}}.
The results are shown in Table~\ref{tab:cifar_reflect}.
We observed that training with one padding method and evaluating using the other resulted in a significant drop in accuracy.
Training using one of these methods randomly for each example resulted in a model which roughly matched the best-case of the two individual cases.

\begin{table}[htp]
\begin{center}
{\renewcommand{\arraystretch}{1.4}
\begin{tabular}{c|c|cc|cc}
    & Natural 
    & \begin{tabular}{c}Random\\ (Zero)\end{tabular}
    & \begin{tabular}{c}Random\\ (Mirror)\end{tabular}
    & \begin{tabular}{c}Grid Search\\ (Zero)\end{tabular}
    & \begin{tabular}{c}Grid Search\\ (Mirror)\end{tabular}\\
\hline                                                                              
Standard Nat
    & 92.62\% & 60.76\% & 66.42\% &  8.08\% &  5.37\% \\
Standard Adv
    & 80.21\% & 59.79\% & 67.12\% &  7.20\% & 12.89\% \\
Aug. A, Zero
    & 90.25\% & 91.09\% & 87.67\% & 59.87\% & 40.55\% \\
Aug. B, Zero
    & 89.55\% & 91.40\% & 87.94\% & 62.42\% & 42.37\% \\
Aug. A, Mirror
    & 92.25\% & 88.43\% & 91.05\% & 41.46\% & 53.95\% \\
Aug. B, Mirror
    & 92.03\% & 88.58\% & 91.34\% & 45.44\% & 57.97\% \\
Aug. A, Both
    & 91.80\% & 90.98\% & 91.28\% & 56.95\% & 52.60\% \\
Aug. B, Both
    & 91.57\% & 91.87\% & 91.11\% & 60.46\% & 56.13\% \\
\end{tabular}}
\end{center}
\caption{
CIFAR10:
The effect of using reflection or zero padding when training a model. The experimental setup matches that of Section~\ref{sec:exp}. Zero padding refers to filling the empty pixels caused by translations and rotations with black. Mirror padding corresponds to using a reflection of the images. "Both" refers to training using both methods and alternating randomly between them for each training example.
}
\label{tab:cifar_reflect}
\end{table}

\end{document}